\documentclass{article}

\usepackage[doublespacing]{setspace}
\usepackage{mathptmx}
\usepackage{times}
\usepackage{amssymb,amsmath} 
\usepackage{mathtools}
\usepackage{amsthm} 
\usepackage[caption=false]{subfig}
\usepackage[pdftex]{graphicx}
\usepackage{algorithm}
\usepackage{algorithmic}
\usepackage{balance}
\usepackage{multirow}
\usepackage{balance}
\usepackage{appendix}
\usepackage[round]{natbib}
\usepackage{hyperref}
\hypersetup{
    colorlinks=true,
    urlcolor=blue,
    citecolor=black,
    linkcolor=black,
}

\theoremstyle{definition}

\newcommand{\funcvar}[1]{\mbox{\tt {#1}}}
\newcommand{\Conf}[0]{\funcvar{Conf}}

\newcommand{\Len}[0]{\funcvar{Len}}
\newcommand{\numtrees}[0]{29 }
\newcommand{\accperc}[0]{70\% }

\usepackage[usenames,dvipsnames]{xcolor}

\usepackage{lineno}
\usepackage{geometry}
\begin{document}
%
\title{\LARGE \bf Semantics-guided Skeletonization of Sweet Cherry Trees for Robotic Pruning}

\author{Alexander You$^1$, Cindy Grimm$^1$, Abhisesh Silwal$^2$, Joseph R. Davidson$^1$
\thanks{This research is supported by the Washington Tree Fruit Research Commission (Award No. TR18103B) and the United States Department of Agriculture-National Institute of Food and Agriculture through the Agriculture and Food Research Initiative, Agricultural Engineering Program (Award No. 2020-67021-31958)}%
\thanks{$^1$Collaborative Robotics and Intelligent Systems (CoRIS) Institute, Oregon State University, Corvallis, OR 97331, USA {\tt\footnotesize \{youa,grimmc, joseph.davidson\}@oregonstate.edu}}%
\thanks{$^2$Field Robotics Center, Carnegie Mellon University, Pittsburgh, PA 15213, USA {\tt\footnotesize{(asilwal@andrew.cmu.edu)}}}
}


%



\maketitle

\begin{abstract}
Dormant pruning for fresh market fruit trees is a relatively unexplored application of agricultural robotics for which few end-to-end systems exist. One of the biggest challenges in creating an autonomous pruning system is the need to reconstruct a model of a tree which is accurate and informative enough to be useful for deciding where to cut. One useful structure for modeling a tree is a skeleton: a 1D, lightweight representation of the geometry and the topology of a tree. This \textit{skeletonization} problem is an important one within the field of computer graphics, and a number of algorithms have been specifically developed for the task of modeling trees. These skeletonization algorithms have largely addressed the problem as a geometric one. In agricultural contexts, however, the parts of the tree have distinct labels, such as the trunk, supporting branches, etc. This labeled structure is important for understanding where to prune. We introduce an algorithm which produces such a labeled skeleton, using the topological and geometric priors associated with these labels to improve our skeletons. We test our skeletonization algorithm on point clouds from \numtrees upright fruiting offshoot (UFO) trees and demonstrate a median accuracy of \accperc with respect to a human-evaluated gold standard. We also make point cloud scans of 82 UFO trees open-source to other researchers. Our work represents a significant first step towards a robust tree modeling framework which can be used in an autonomous pruning system.



\end{abstract}

\textbf{Keywords}: Tree modeling, Skeletonization, Dormant pruning, Robotic pruning, Agricultural robotics

\begin{figure*}[bt]
	\centering
\includegraphics[width=0.95\linewidth]{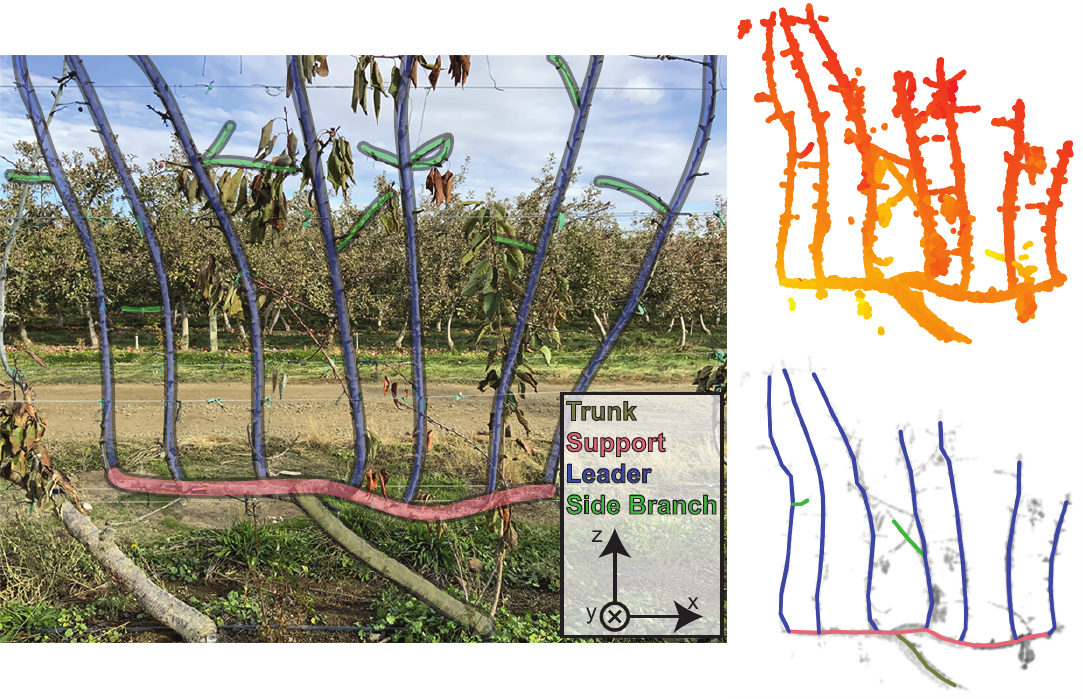}
\caption{Left: A bare UFO cherry tree, along with the labelled parts of the tree and a coordinate reference frame. Top-right: Tree point cloud, colorized by y-coordinate. Bottom-right: Resulting skeleton overlaid on the point cloud.}
\label{fig:graphicalabstract}
    \vspace{-2ex}
\end{figure*}

\section{Introduction}
Production of high-value tree fruit crops --- such as fresh market apples, pears, and sweet cherries --- requires a large, semi-skilled workforce for short, intensive periods during the year. After harvesting, the most labor-intensive orchard activity is dormant season pruning~\citep{verbiest2020automation}, a highly repetitive and potentially dangerous process that involves using hand tools (e.g. shears and loppers) and ladders to selectively remove branches that are too old, diseased, or are otherwise unproductive. Pruning is a critical perennial operation required to maintain tree health and produce high yields of quality fruit. While there has been some work on using robotic systems to prune grapevines~\citep{botterill2017robot,VisRob}, the technology is still in its infancy and has received relatively little attention compared to robotic harvesting. As the tree fruit industry continues to face the challenge of rising labor costs and growing uncertainty about the future availability of seasonal workers, the development of robotic pruning technology will play a critical role in its long-term sustainability.

A successful end-to-end pruning system must be capable of accurately sensing the environment, creating a robust internal representation of its environment, using this representation to decide where to cut, planning a collision-free motion, and finally moving a cutting implement to the desired cut points. Each of these steps is itself a challenging problem, requiring state-of-the-art performance from a variety of robotics and computer science subfields. In our previous work~\citep{you2020}, we demonstrated that fast, efficient, closed-loop planning to reach manually-specified cut points in the environment is achievable. This paper focuses on developing the steps of the pipeline which can automatically determine those cut points. The start of this process is to capture a point cloud of the tree and turn it into a compact, but descriptive, representation of that tree suitable for planning. An appropriate such representation is a \textit{skeleton}, i.e. a set of points in $\mathbb{R}^3$ and connections between them which describe the overall topology of the structure. This process, known as \textit{skeletonization}, has been widely studied in the field of computer graphics, and has also been previously applied to model organic structures such as trees and other plants. However, in the agricultural domain, because many applications rely on an understanding of the semantic structure of the trees (such as the one shown in Figure~\ref{fig:graphicalabstract}), our skeletonization algorithm must be capable of recovering not just the geometry but also the corresponding semantic labels. For pruning in particular, it is essential that this structure be correct, as missing, adding, or mislabeling branches can lead to task failure and even structural damage caused by incorrect cuts. Fortunately, the semantic structure also introduces geometric and structural constraints which can be used to guide and cross-validate the skeletonization process, ultimately resulting in more accurate skeletons.

Our primary contribution is the development of a skeletonization algorithm that uses basic assumptions about tree structure to output a skeleton with labelled edges corresponding to that structure, as shown in Figure~\ref{fig:graphicalabstract}. Our algorithm takes advantage of developments in machine learning and probabilistic search algorithms to evaluate the plausibility of candidate skeletons and select the best ones. We validate our algorithm by asking human experts to evaluate the correctness of the labels and skeleton.  We analyze \numtrees trees, with the initial results showing that a skeleton produced by our algorithm resulted in an overall accuracy of \accperc when compared against a corrected skeleton.  These results demonstrate that, with a minimal degree of human intervention, a quick approach to automated skeletonization is feasible. We also make our collection of 82 scanned UFO trees available for other researchers \href{https://drive.google.com/drive/folders/1V2XNrYTp715YA0iOg8Ewn-oLq1Bha6m_}{at this link} (see Section~\ref{sec:datacollection} for the full URL).

Our paper is organized as follows. We first discuss the tree structure and our general assumptions about the orchard system selected for this study, as well as how these assumptions should be reflected in the resulting skeleton (Section~\ref{sec:problem}). We then introduce our algorithmic framework for processing a point cloud into a labelled skeleton (Section~\ref{sec:methodology}). This algorithm is broken into two steps: i) A pre-processing step which reduces the point cloud to a more manageable superpoint-based graph, followed by ii) a population-based search on this graph which explores many possible candidate skeletons, growing them one labelled edge at a time. We describe how we collected the point cloud data from the orchard in Section~\ref{sec:experiment}, and then discuss evaluation of the accuracy of the skeletonization algorithm on these point clouds (using human labelling as the gold standard) in Section~\ref{sec:results}. We conclude with a discussion of future avenues for research.


\section{Related work}

Few end-to-end robotic pruning systems have been successfully demonstrated; the main systems of which we are aware are grapevine pruning robots from~\cite{VisRob} and~\cite{botterill2017robot} as well as a rose pruning robot from~\cite{cuevas2020real}. \cite{botterill2017robot} detail the process by which they combine stereo image pairs along with an image segmentation map to form the 3D model from which they extract cut points; this process differs slightly from our own since we assume that the input is already a 3D point cloud. \cite{he2018sensing} cover a wide variety of existing research on pruning for apple trees, much of which is focused on methods of sensing the environment. While most of this research was focused on modeling rather than using these models to decide cut points, some research such as~\cite{karkee2015method} and~\cite{tabb2017robotic} constructs skeletons from stereo image pairs and focuses on reconstructing important traits (i.e. phenotypes) such as branch length and thickness. \cite{karkee2015method} explicitly mention that these traits could be used to develop pruning rules, showcasing preliminary success on tall spindle apple trees~\citep{karkee2014identification}. Our previous work demonstrated an efficient framework for executing manually-specified pruning cuts~\citep{you2020}. Our goal here is to develop a skeletonization algorithm which, using the output skeleton and branch classifications, can be used to automatically determine desired cut poses.

The problem of skeletonization for describing the topology of objects such as grid structures, point clouds, and meshes is an area of active research in the field of computer graphics. Two recent reviews cover a wide variety of modern 3D skeletonization techniques on data types such as voxel representations, unorganized point clouds, and meshes~\citep{tagliasacchi20163d,saha2016survey}. Our particular area of interest is skeletonization algorithms for unorganized 3D point cloud data of trees in outdoors environments, many of which originate from areas such as agriculture and forestry. There are many challenges associated with obtaining accurate point clouds of outdoor environments. \cite{akbar2016novel} focus on point clouds of tree branches for branch reconstruction, citing issues such as occlusions, lighting variations, background noise, and inherent sensor noise. An ideal skeletonization algorithm should be able to overcome these deficiencies. Historical approaches to skeletonizing these types of point clouds can be roughly split into graph-based methods and iterative optimization-based methods.

Graph-based skeletonization methods work by creating a dense graph and applying a set of decisions to thin the graph, leaving behind the most essential topological components. Many such algorithms operate by converting a point cloud into an octree structure and defining a series of graph-based rules to eliminate local cycles in the graph and thin the structure. The CAMPINO algorithm~\citep{bucksch2008campino} forms an octree and uses a set of rules to identify local cycles with specific structures and reduce them into acyclic structures. SkelTre~\citep{bucksch2010skeltre} operates by finding eligible pairs of connected edges and reducing them into a single edge. It also includes a basic point-distance criterion for filtering out false connections between adjacent cells of the octree. An extension to SkelTre~\citep{zhang20163d} constructs a skeleton by first splitting up the octree into individual branch segments. This splitting is done via a flood fill procedure which tries to detect when the octree has forked off into a new branch by counting the voxels added in each direction at each step. It then runs the SkelTre algorithm on each branch segment and joins the segments together to obtain the skeleton. The advantage of these graph-based algorithms lie primarily in their speed, with linear-time processing of the graph~\citep{bucksch2010skeltre}. However, because they do not consider known priors about the tree geometry, this can lead to issues such as disconnected branches or unusual branch shapes. Other non-voxel, graph-based methods attempt to incorporate tree geometry heuristics in forming the skeleton. \cite{xu2007knowledge} perform an initial skeletonization which may contain disconnected components and attempt to connect these components to the main skeleton via a breadth-first search on the main skeleton for promising connection points. \cite{livny2010automatic} make assumptions about the geometry of the tree, e.g. that branches should be relatively smooth, and turn these assumptions into an optimization constraint which is used to smooth out a dense graph representation of the original point cloud. Out of all previous approaches, this last approach is perhaps the most similar to our own.

Another set of methods revolves around embedding a set of skeleton points inside of a point cloud through iterative optimizations. One such well-known method is the $L_1$-medial skeleton~\citep{huang2013l1}, which distributes a set of skeleton points within a point cloud via repulsive forces such that each point represents the center of a maximal ball contained within the point cloud. This method is fairly robust to noise and outliers, but it is not guaranteed to preserve the tree topology of a point cloud. Others have built on top of it to improve the resulting skeletons for real trees. \cite{mei20173d} use  the $L_1$-median algorithm to extract a coarse skeleton; they then use a minimum spanning tree-based technique to connect disconnected branches, followed by a smoothing technique using a method of Laplacian contraction developed by~\cite{cao2010point}. Previous work from~\cite{wang2016local} also showed that this same method of Laplacian contraction could be used to iteratively repair point clouds with missing data by using branch directionality information to push the skeleton points towards areas of missing data. \cite{fu2020tree} further build upon the $L1$-medial skeleton method by introducing a constraint based off the knowledge that branches are roughly cylindrical, allowing for more robust centering of the skeleton under missing data.

One potentially promising avenue for improving the robustness of skeletonization algorithms is to incorporate machine learning methods, which have shown robustness on visual tasks with which heuristics have had limited success. In 2D, ~\cite{botterill2013finding} apply support vector machines (SVMs), a more traditional machine learning method, on 2D edges to determine whether two branch segments are connected based off a vector of geometric characteristics. Convolutional neural networks (CNNs) are naturally designed to operate on organized grids of data. This approach has been used to train a binary classifier to label skeleton pixels~\citep{shen2017deepskeleton}. Specifically, the network is trained to recognize pixel scales via embedded circles in order to produce accurate skeleton pixel maps. \cite{cuevas2020segmentation} use an RGB-D camera and applies traditional skeletonization methods to the output of a semantic segmentation network along with a corresponding depth map to construct 3D skeletons of rose plants. However, significantly less research exists on applying deep networks to unorganized point cloud data. \cite{demir2019skelneton} applied PointNet++~\citep{qi2017pointnet++}, a neural network architecture that directly uses point cloud inputs without needing rasterization, to try to classify points as skeletal and non-skeletal. The skeletons produced by this approach were of limited success, often forming discontinuities and failing to include long and narrow shapes in the skeleton. In this paper, we propose a simple use of a 2D CNN that assists in robustly determining connections between skeleton points.

\section{Problem description}
\label{sec:problem}
In this section we describe the semantic tree labeling problem. First, we describe the structure of an Upright Fruiting Offshoot (UFO) orchard system, the type of tree we scanned for our experimental evaluation (Section~\ref{sec:treesescription}). Next, we formally define a labeled tree skeleton and introduce the terminology used in the remainder of the paper (Section~\ref{sec:term}). Finally, we define the specific geometric and topological constraints for the UFO trees (Section~\ref{sec:ufo_constraints}). We include a discussion on how this setup can apply generally to other types of trees in Section~\ref{sec:discussion}. 

\subsection{UFO tree description}
\label{sec:treesescription}


For this work, we focus on the Upright Fruiting Offshoot (UFO) orchard system, which has been recently adopted by some Pacific Northwest sweet cherry growers due to its increased fruit yield and amenability to automation. We discuss how to extend this work to other types of tree structures in Section~\ref{sec:discussion}. A typical UFO tree and its structure are shown in Figure~\ref{fig:graphicalabstract}. Their most defining feature is the \textbf{leader} branches, which grow vertically to a height of approximately 3m and are supported by horizontal trellis wires. There are typically 7 to 9 of these leaders on an individual tree spaced 15-45 cm apart. The leaders are where the tree bears cherries during the harvesting season~\citep{long2015cherry}, and the goal of dormant pruning for these trees is to remove \textbf{side branches} of a sufficient length that grow off of the leader branches. The vertical leaders grow up from a horizontal \textbf{support} branch (or cordon) at a height of approximately 60 cm above grade. The two horizontal support branches are connected to the ground via the \textbf{trunk}. The leaders, side branches, support, and trunk form the primary structure of the tree which we wish to extract.

\subsection{Skeletonization: formal definition}
\label{sec:term}


In this section we formally define skeletonization as a graph construction problem. The input to our algorithm is a point cloud $P = \{p_i\}$ with $p_i \in \mathbb{R}^3$. Our output is a \textbf{labelled skeleton} $S = (N, E, L)$, defined on a set of nodes $N$:

\begin{itemize}
    \item $N = \{n_i\}$ represents a finite set of points in $\mathbb{R}^3$ which we call \textit{superpoints}. $N$ is a downsampling of the point cloud $P$ where the points $n_i$ are averages of point locations in $P$.
    
    \item $E = \{e_j\}$ represents a set of directed edges which connect unique superpoints in $N$, i.e. $e_j = (n_{j_1}, n_{j_2})$, $n_{j_{1,2}} \in N$, and $n_{j_1} \neq n_{j_2}$.
    
    \item $L : e_j \in E \rightarrow \{{\tt Trunk, Support, Leader, Side Branch, None}\}$ is a mapping of each edge in $E$ to a unique label. This label set is specific to the UFO cherry tree structure.
\end{itemize}

Because of the one-to-one correspondence of the edges $E$ and the labels $L$, we will write as shorthand that $(e, l) \in S$ if there exists some $j$ such that $e_j = e \in E$ and $L(e) = l$.

In order to form a tree structure, we require that the undirected graph $(N, E)$ forms a tree, i.e. a connected acyclic graph. Furthermore, all of the edges in $E$ must be directed to form an out-tree, i.e. there exists exactly one node $n_B$, called a \textit{base node}, which acts as a source node. In other words, given a node $n \in N$ such that there is at least one edge $e \in E$ with $n \in e$, there exists a unique set of edges in $E$ forming a path which starts at $n_B$ and ends at $n$. This also implies that every edge has exactly one predecessor edge, with the exception of the edge containing $n_B$, which will not have a predecessor edge.



\subsection{Label-based constraints and assumptions for the UFO tree}
\label{sec:ufo_constraints}

Unlike traditional skeletonization algorithms, which make few assumptions about the object being skeletonized, our goal is to utilize our prior knowledge of the UFO tree structure to produce a high-quality, accurate skeleton. In particular, in order to guide the skeletonization process, we utilize two sets of label-based constraints: topological constraints (Section~\ref{sec:topological}) and geometric constraints (Section~\ref{sec:geomcon}).

\subsubsection{Label-based topological constraints}
\label{sec:topological}

\begin{figure}[t]
	\centering
\includegraphics[width = \columnwidth]{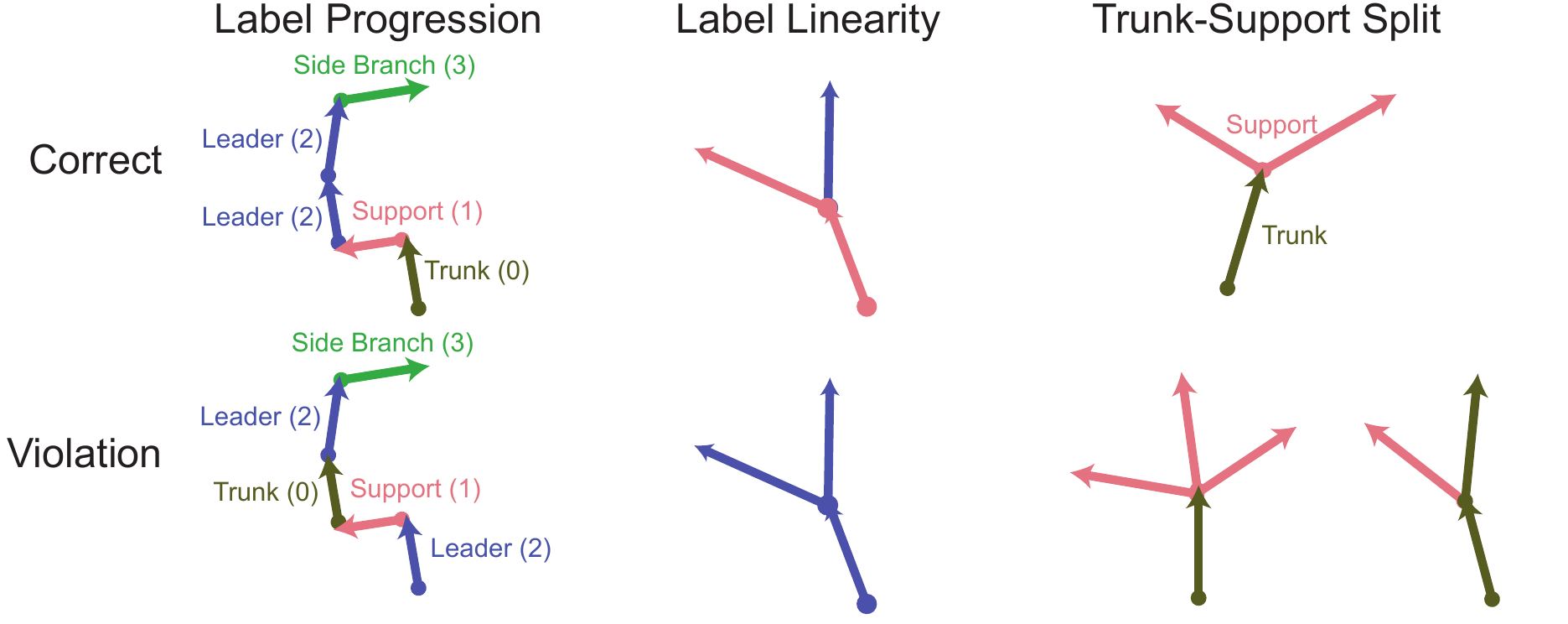}
\caption{An illustration of the label-based topological constraints on the skeleton. Label Progression: Edge labels must progress in a given order. Label Linearity: There cannot be a Y-junction in which all of the edges have the same label. Trunk-Support Split: A trunk edge cannot be succeeded by more than two support edges, and a non-trunk edge may not connect to a superpoint node which is both preceded and succeeded by a trunk edge.}
    \label{fig:topologicalconstraints}
\end{figure}

In addition to the topological constraints on the directions of edges in $E$, we also impose topology-based constraints on the labels assigned to these edges. As noted in Section \ref{sec:treesescription}, the structure of the tree dictates that side branches grow out from the leaders, leaders grow out from the support, the support will grow out from the trunk, and there are at most two supports. Our first label constraint, \textbf{label progression}, ensures that the edge labels progress in this specified order to prevent erroneous label assignments, e.g. a support growing out of a leader branch. The second constraint, \textbf{label linearity}, prevents Y-shaped junctions from all having the same label. The final constraint, the \textbf{trunk-support split}, dictates how the support branches may grow out of the trunk. Specifically, the support branch must connect at the very top of the trunk (i.e. it cannot connect ``halfway'' down a trunk), and there cannot be more than two support edges connected to the trunk edge. These constraints are illustrated in Figure~\ref{fig:topologicalconstraints}.

More formally, we define the topological edge label constraints as follows:

\begin{itemize}
    \item \textbf{Label progression}: Let $e_s$ be a successor edge of $e_p$.  Define a function $Order(l)$ which returns $0$ if $l=\funcvar{Trunk}$, $1$ for $l=\funcvar{Support}$, $2$ for $l=\funcvar{Leader}$, and $3$ for $l=\funcvar{SideBranch}$. Then $Order(L(e_p)) \leq Order(L(e_s))$. 
    
    \item \textbf{Label linearity}:  Suppose $e_{s_1}$ and $e_{s_2}$ are successors to the edge $e_p$ with $e_{s_1} \neq e_{s_2}$. Then $L(e_{s_1})$, $L(e_{s_2})$, and $L(e_{p})$ cannot all be the same label.
    
    \item \textbf{Trunk-support split}:
    Let $e_p$ be an edge with $L(e_p) = \funcvar{Trunk}$. Let $Succ(e_p)$ be the set of successor edges of $e_p$. If there exists $l \in L(Succ(e_p))$ such that $l \neq \funcvar{Trunk}$, then $\funcvar{Trunk} \notin L(Succ(e_p))$, and at most two of these edges may have $\funcvar{Support}$ labels.
\end{itemize}

\subsubsection{Label-based geometric constraints}
\label{sec:geomcon}


In addition to the label topology constraints from before, we also utilize general knowledge about UFO trees to evaluate the likelihood of an edge-label assignment. First, trees branches generally grow fairly straight unless at a branching point, a rule which applies to a wide variety of trees. For UFO trees specifically, we also know that the supports are mostly horizontal and the vertical leaders are vertical. To define the growth direction, we take advantage of the fact that our tree is mostly planar and lies in the $XZ$ plane ($X$ is across the width of the tree, while $Z$ is up in the world). Due to variations in the shape of UFO trunks and side branches, we do not apply growth direction constraints for these labels.

Define $\overrightarrow{e}$ as the vector formed by the edge $e$, i.e. $\overrightarrow{e} = n_{j_2} - n_{j_1}$. We formally quantify ``straightness'' via the \textbf{turn angle}, as well as an edge's \textbf{growth direction}:

\begin{itemize}
    \item \textbf{Turn angle}: For an edge $e_s$ and its predecessor $e_p$, the branch turn angle is the complement of the angle formed by the edges:
    
    \begin{equation}
    \label{eq:anglechange}
    \angle(\overrightarrow{e_s}, \overrightarrow{e_p}) = \arccos\left(\frac{ \overrightarrow{e_s}  \cdot \overrightarrow{e_p}}{\|\overrightarrow{e_s}\| \|\overrightarrow{e_{p}}\|}\right)
    \end{equation}
    
    If two adjacent edges have the same label then their turn angles should be low, since  branches tend to be locally smooth unless located at a branching point.
    
    \item \textbf{Growth direction}: For an edge $e$, its growth angle is the angle formed by the edge in the $XZ$-plane with respect to the $X$-axis:
    
    \begin{equation}
    \label{eq:angle}
    \funcvar{GrowAngle}(e) = \arctan\left(\frac{|\overrightarrow{e} \cdot \langle0, 0, 1\rangle|}{|\overrightarrow{e} \cdot \langle1, 0, 0\rangle|}\right)
    \end{equation}

    Leader edges should be close to vertical ($\pi/2$) and support edges should be close to horizontal ($0$).
\end{itemize}

These geometric constraints represent ``soft'' constraints, as it may be OK for them to be temporarily violated, such as a leader branch growing out in the y-direction where it connects to the support. We discuss how these constraints are incorporated into the skeletonization process in Section~\ref{sec:objfunc}.

\section{Methodology}
\label{sec:methodology}

\begin{figure}[p]
	\centering
\includegraphics[width = \columnwidth]{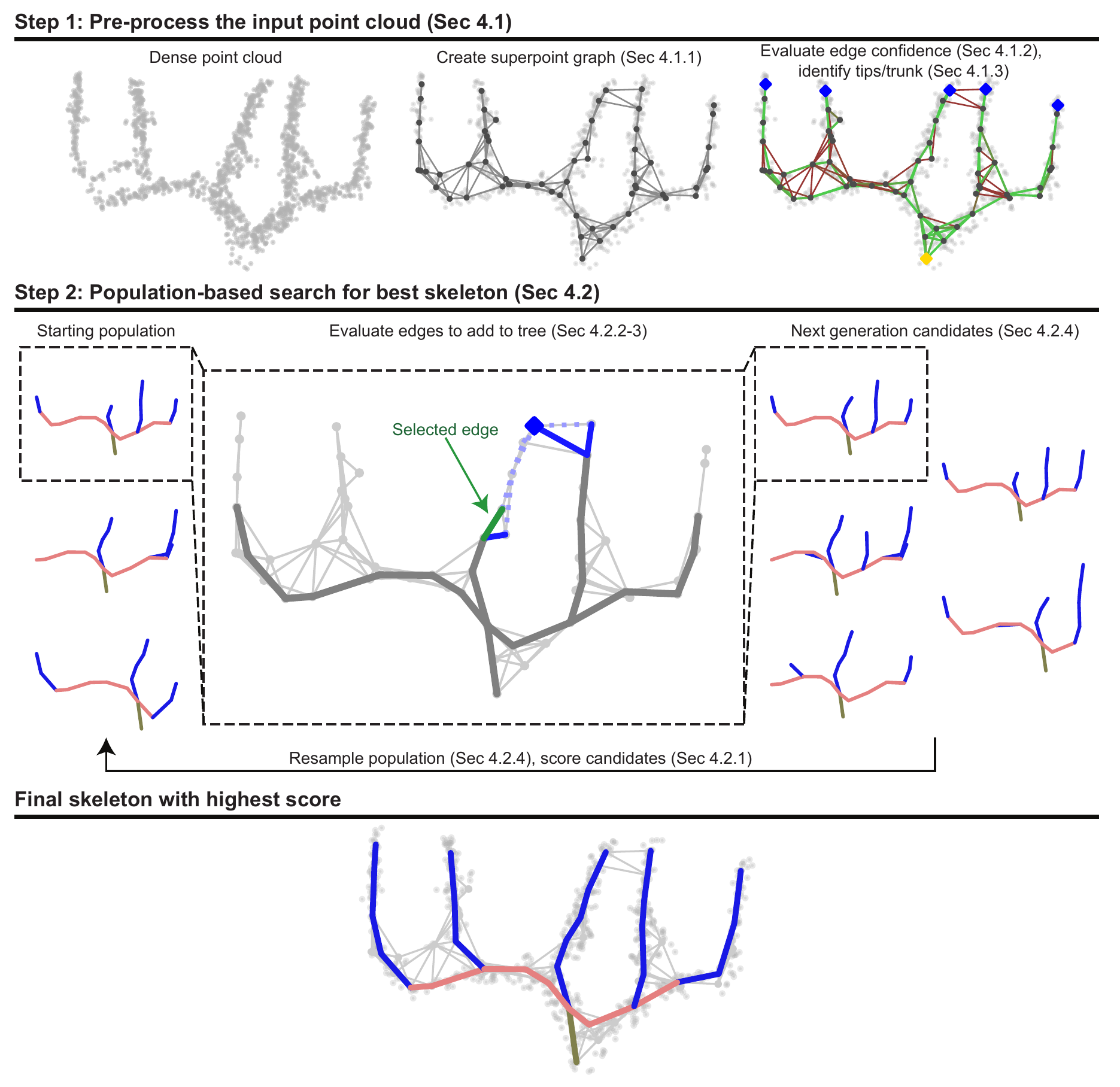}
\caption{Our pipeline for constructing a labelled skeleton from the input point cloud. Top row, pre-processing: The point cloud is downsampled to superpoints which are connected to their nearby neighbors. The base node (yellow dot) and tip nodes (blue) are identified and the quality of each edge evaluated. Middle row, population-based search: A population of skeletons is iteratively grown, adding one edge at a time to each candidate. Candidates are scored and the best continue to the next iteration.}
    \label{fig:generaloverview}
\end{figure}

The two-step tree skeletonization algorithm is outlined in Figure~\ref{fig:generaloverview}. The first step is a \textit{pre-processing} phase which downsamples the pointcloud to produce a dense, nearest neighbor graph (Section~\ref{sec:preprocessing}). The second phase is a \textit{population-based search} which iteratively constructs possible candidate skeletons one edge-label pair at a time, growing the skeleton from the tree's base towards a set of ``tip'' superpoints representing the tops of the vertical leaders (Section~\ref{sec:population}). For reference, at the end of the section, Table~\ref{table:params} lists the tunable parameters that control the behavior of the algorithm and Table~\ref{table:notation} summarizes all of the notation used.

\subsection{Point cloud pre-processing}
\label{sec:preprocessing}

In the pre-processing phase, the goal is to simplify the point cloud into a graph structure suitable for the optimization phase. The goal is to reduce the total number of points to a manageable size while still ensuring that the edges ``bridge'' any gaps in the point cloud.

This process consists of three main steps. First, the point cloud is downsampled into superpoints (Section~\ref{sec:methodology:sp}). Second, we define a potential set of graph edges (using nearest neighbors) and evaluate each edge on its likeliness to belong to the skeleton (Section~\ref{sec:edgeeval}). Finally, we use the graph and evaluations to determine which superpoints represent possible tips of the vertical leaders (Section~\ref{sec:tip_and_trunk}).

\subsubsection{Superpoint selection}
\label{sec:methodology:sp}

Because the point clouds we obtained typically consisted of about 300,000 points each, reducing the density of the data is essential for efficient skeletonization. As previously noted, voxelization is a common sparsification technique which reduces a dense point cloud into a smaller set of occupied cells in a grid space. We use a method similar to voxelization to reduce subsets of the point cloud to a single super point, but without the rigid grid structure. Specifically, given a radius $r_{super}$ we cover the original point cloud $\mathbf{P}$ with slightly overlapping spheres of radius $r_{super}$. We call the centers of each of these spheres \textit{superpoints}, which then form our set of skeleton nodes $N$. The two underlying assumptions which motivate this approach and influence the choice of $r_{super}$ are:

\begin{itemize}
    \item $r_{super}$ is bigger than the radius of the branches, ensuring that branch segments are not split width-wise across a superpoint.
    \item $r_{super}$ is smaller than the average distance between leaders, ensuring that superpoints do not span multiple leaders.
\end{itemize}

The superpoint cover is formed as follows. While the point cloud $P$ is not fully covered, we randomly select an uncovered point $p_i \in P$ and identify the set of points 
$\{p_j: \left|p_j - p_i \right| \leq r_{super}\}$. We then set the corresponding superpoint $n_i$ to be at the mean of the points in $\{p_j\}$ and add $n_i$ to the set of superpoints $N$. We then mark all of $\{p_j\}$ as being covered. This process continues until all points in $P$ are covered. Note that we allow the spheres to overlap; a point $p_j$ may be within $r_{super}$ of more than one node $n_i$. With our choice of $r_{super} = 0.10$, superpoint graphs typically consisted of between 150 to 300 superpoints.

\subsubsection{Edge evaluation}
\label{sec:edgeeval}

\begin{figure}[tb]
	\centering
\includegraphics[width=90mm]{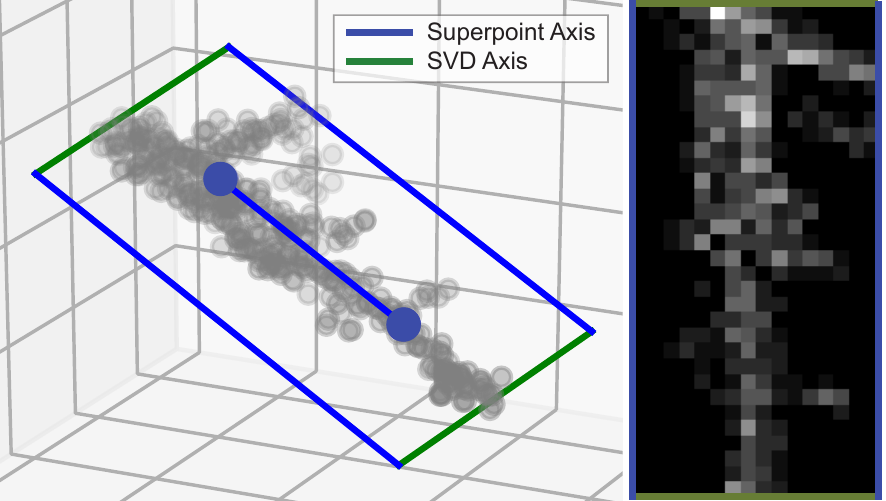}
\caption{Evaluating a potential branch. Left: The point cloud is projected orthogonal to the edge in question by fitting a plane to the point cloud. Right: The resulting image is evaluated by a CNN to determine if it is branch-like.}
\label{fig:project}
\end{figure}

After the superpoints $N$ are selected, the next step is to propose a candidate set of connecting edges and evaluate their likeliness for being part of the skeleton. Point clouds suffer from a variety of issues such as noise, gaps, and density variations which make developing a reliable heuristic for connectedness difficult. However, our observation is that \textit{visually} determining if an edge exists is not too difficult. This leads to the idea of using a convolutional neural (CNN) approach to score potential edges, as CNNs have shown respectable performance on other visual classification tasks. We first produce a dense set of edges $\overline{E}$ by adding an edge between any two superpoints that are within $2r_{super}$ of each other. We then project the point cloud associated with the two superpoints into an image and evaluate the image to determine if it represents a valid branch (see Figure~\ref{fig:project}).

Specifically, let $e = (n_{i}, n_{j})$ be the proposed superpoint edge, and let $\{p_k\} \subset P$ be the set of points within $r_{super}$ of $n_i$ or $n_j$. We first transform $\{p_k\}$ into a new coordinate frame with the following properties:

\begin{itemize}
    \item The origin is coincident with the midpoint of $e$.
    \item The $x$-axis is coincident with $\overrightarrow{e}$.
    \item The $z$-axis is coincident with the \textit{least} significant component of $\{p_k\}$ when decomposed via SVD, orthogonalized with respect to $\overrightarrow{e}$.
\end{itemize}

After transforming the points, we drop the $z$ component of the points and histogram the 2D points into a $32 \times 16$ bucket grid. Afterwards, we normalize the values in the histogram by the maximum value to obtain a rasterized greyscale image. We then feed this image, along with a vector containing the edge's origin and growth direction, into a CNN which outputs a number $\in [0,1]$ representing the edge validity. This score is used in the following sections to guide the identification of edges which form the main structure of the tree. 

To train the network, we generated 15000 edge images by randomly sampling edges from a training set of point clouds, classifying them as either true connections or false connections. 700 of these edges were manually labelled, while the remaining 14300 were obtained via data augmentation by randomly downsampling the point clouds used to generate the edges, allowing the classifier to work well on a variety of point densities. More details on the neural network are given in Appendix \ref{sec:nn}.




\subsubsection{Tip and trunk identification}
\label{sec:tip_and_trunk}

\begin{figure}[tb]
\centering
\includegraphics[width = 9cm]{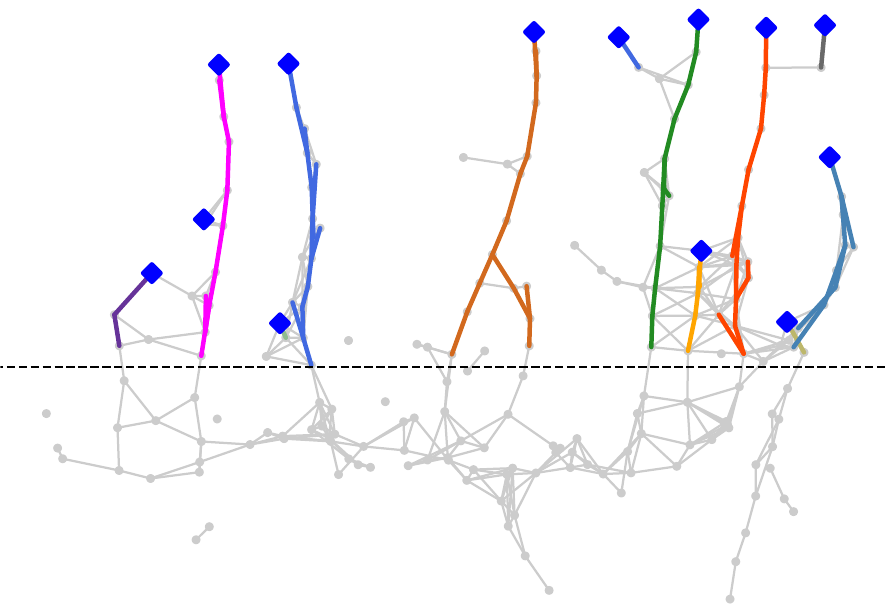}
\caption{Tip superpoint candidates are selected by constructing a confidence-weighted minimum spanning tree from the dense superpoint graph. After discarding edges from the MST which are horizontal or have low confidence, tips are identified within each connected component as the superpoint with the largest $z$-coordinate.}
\label{fig:tipnodes}
\end{figure}

Our growth-based strategy requires us to identify a set of tip superpoints $\{n_t\}  \subseteq N$ which represent the tops of the vertical leaders. The final skeleton is ultimately formed by growing a skeleton starting from the base of the tree towards these tips. However, extracting a set of candidate tips from the dense superpoint graph $(N, \overline{E})$ can be difficult when leaders are located close to other branches or noisy superpoints. Figure~\ref{fig:tipnodes} illustrates our process for identifying potential tip nodes. We essentially build a minimum spanning forest across the entire graph which attempts to preserve high-confidence edges, then ``chop off'' the bottom part of the tree. From each remaining connected component, we identify the highest superpoint in each cluster as a potential tip.

More formally, we first create a copy of the dense edges $\overline{E}$, but remove all edges with a confidence value below a predetermined threshold $\alpha_{conf}$. We then run Kruskal's algorithm to obtain a minimum spanning forest $E^*$ on the modified dense set of edges, where the weight for an edge $e$ is $\Len(e)(1 - \Conf(e))$. This weighting favors including high-confidences edges in the forest. Once we form the minimum spanning forest, we further throw out all edges in $E^*$ which meet at least one of the following conditions:

\begin{itemize}
    \item \textbf{Horizontal branch}: The edge is not sufficiently vertical, i.e. $\funcvar{GrowAngle}(e) < \frac{\pi}{4}$.
    \item \textbf{Bottom of tree}: The $z$-coordinate of one of the edge's endpoints is less than $z_{max} - \alpha_{tip}(z_{max} - z_{min})$, where $\alpha_{tip}$ is a predetermined height threshold value between 0 and 1 (we choose it to be 0.6).
\end{itemize}

Once these edges are removed, we identify the connected components in the subgraph. For each connected component, we find the superpoint with the maximum $z$-value and add it to our set of candidate tips $\{N_t\}$. Although some of these tips may be spurious and not correspond to a real leader tip, they will eventually be filtered out by the search algorithm.

Our algorithm also requires us to identify a base superpoint $n_B \in N$ representing the location where the trunk of the tree meets the ground. Presently, because of the noise in the point cloud in the ground plane, we select this node manually. Section~\ref{sec:discussion} discusses possible methods for automatic identification. 


\subsection{Population-based search}
\label{sec:population}

The pre-processing step produces a dense graph $(N, \overline{E})$ and identifies a base node $n_B$ and a set of candidate tip superpoints $\{n_t\} \subset N$, as well as confidence scores for each edge in the mapping $\Conf: \overline{E} \rightarrow [0, 1]$. The goal of the population-based search is to take these inputs and run an optimization to ``grow'' a tree that satisfies our topological criteria while including high-confidence edges and maximizing our geometric criteria. The key is that instead of growing just one skeleton candidate, we keep track of \textit{many} hypothetical skeletons. Doing so allows us to prevent a situation in which a single bad assignment leads to a suboptimal skeleton, since early skeleton modifications can significantly impact the skeleton's growth later in the process. This population-based approach is inspired by fitness-based optimization methods such as evolutionary algorithms.

Figure~\ref{fig:generaloverview}, middle, outlines the iterative optimization process. The algorithm starts with a set of $K$ empty skeleton candidates. The core of the algorithm is a loop which attempts to identify the main structural elements, i.e. the trunk, support, and leaders:

\begin{enumerate}
    \item Score each of the $K$ skeleton candidates in the population, and update the best tree across all iterations. (Section~\ref{sec:objfunc})
    \item For each skeleton candidate, pick a random tip superpoint and determine possible edges which could be added to the skeleton and which grow towards the tip, along with possible labels. (Section~\ref{sec:edgeproposal})
    \item Rank the possible edge-label propositions within each skeleton candidate. (Section~\ref{sec:potential})
    \item Compute a weight for each possible next-generation skeleton. Then use these weights to create the new population of $K$ candidates for the next iteration via a weighted random sampling. (Section~\ref{sec:pop_sample}) 
\end{enumerate}

Once this iterative process has terminated, we run a post-processing step to find side branches. (Section~\ref{sec:sidebranches})

\subsubsection{Skeleton evaluation objective function}
\label{sec:objfunc}

This section defines the objective function, the skeleton evaluation function $\funcvar{SkelScore}(S) \rightarrow \mathbb{R}$, which our search process is designed to maximize. $\funcvar{SkelScore}$ is a measure of skeleton quality which scores a potential skeleton based on the edge quality, how straight the branches are, and if the supports and leaders are oriented correctly. Because we favor skeletons that cover as much of the graph as possible, the score {\em increases} with each edge added (provided that we are adding ``good'' edges). In optimization terms, our approach is one of reward maximization, as opposed to cost minimization. Although cost minimization problems are generally easier to solve for~\citep{stern2014max}, our need to cover as much of the tree as possible means a maximization approach is more suitable.

$\funcvar{SkelScore}$ is simply defined as the sum of the rewards of each edge-label pair in the skeleton:

\begin{equation}
    \funcvar{SkelScore}(S) = \sum_{e \in E} \funcvar{Reward}(e,L)
    \label{eq:skelscore}
\end{equation}

\begin{equation}
    \funcvar{Reward}(e, l) = \funcvar{EdgeScore}(e) - \funcvar{TurnPenalty}(e,\, Pred(e),\, L) - \funcvar{GrowPenalty}(e, L)
\end{equation}

The $\funcvar{Reward}$ function is, in turn, comprised of three components. $\funcvar{EdgeScore}$ is based off of our confidence score for the edge (Section~\ref{sec:edgeeval}) and a preference for longer edges. The $\funcvar{TurnPenalty}$ and $\funcvar{GrowPenalty}$ components are penalties based off of our tree geometry criteria (Section \ref{sec:geomcon}). 

\noindent $\funcvar{EdgeScore}$ weights scores by the corresponding edge lengths and their confidence values, returning a negative score if the confidence value is below the given confidence threshold $\alpha_{conf}$:
    
    \begin{equation}
    \label{eq:edgescore}
    \funcvar{EdgeScore}(e) = \funcvar{Len}(e) \left(1 - \frac{1 - \funcvar{Conf}(e)}{1 - \alpha_{conf}}\right)
    \end{equation}
    
\noindent $\funcvar{TurnPenalty}$ penalizes adjacent edges that have the same label but bend more than a threshold $\theta_{angleMin}$. Let $e_s$ and $e_p$ be an edge and its predecessor (if $E$ is fixed, we denote the predecessor of $e_s$ in $E$ as $Pred(e_s)$), and $c_{Turn}, p_{Turn}$ be constants:
    
    \begin{equation}
    \label{eq:turnpenalty}
    \funcvar{TurnPenalty}(e_s,\, e_p,\, L) = \begin{cases*}
        0 & if $L(e_s) \neq L(e_p)$ and $L(e_s) \neq \funcvar{None}$ \\
        0 & if $\angle\left(e_s, e_p \right) \leq \theta_{angleMin}$ \\
        c_{Turn}\left(\angle\left(e_s, e_p \right) - \theta_{TurnMin}\right)^{p_{Turn}} & otherwise
    \end{cases*}
    \end{equation}
    
\noindent $\funcvar{GrowPenalty}$ penalizes support and leader edges that are not horizontal and vertical respectively by more than a threshold $\theta_{GrowMin}$. Let $c_{Grow}, p_{Grow}$ be constants, and $\Delta\funcvar{Target}$ specify the deviation from the expected growth angle for the edge:
    
\begin{equation}
\label{eq:growpenalty}
    \funcvar{GrowPenalty}(e, L) = \begin{cases*}
    0 & if $l \notin \{\funcvar{Leader}, \funcvar{Support}\}$\\
    0 & if $\Delta \funcvar{Target}(e, L) \leq \theta_{GrowMin}$ \\
    c_{Grow}\left(\Delta \funcvar{Target}(e, L) - \theta_{GrowMin}\right)^{p_{Grow}} & otherwise
    \end{cases*}
    \end{equation}
    
    \begin{equation}
        \Delta \funcvar{Target}(e, L) = \begin{cases*}
            \funcvar{GrowAngle}(e) & if $L(e) = \funcvar{Support}$ \\
            \frac{\pi}{2} - \funcvar{GrowAngle}(e) & if $L(e) = \funcvar{Leader}$
        \end{cases*}
    \end{equation}

\subsubsection{Determining eligible edge-label pairs}
\label{sec:edgeproposal}

\begin{figure}[bt]
	\centering
\includegraphics[width = \columnwidth]{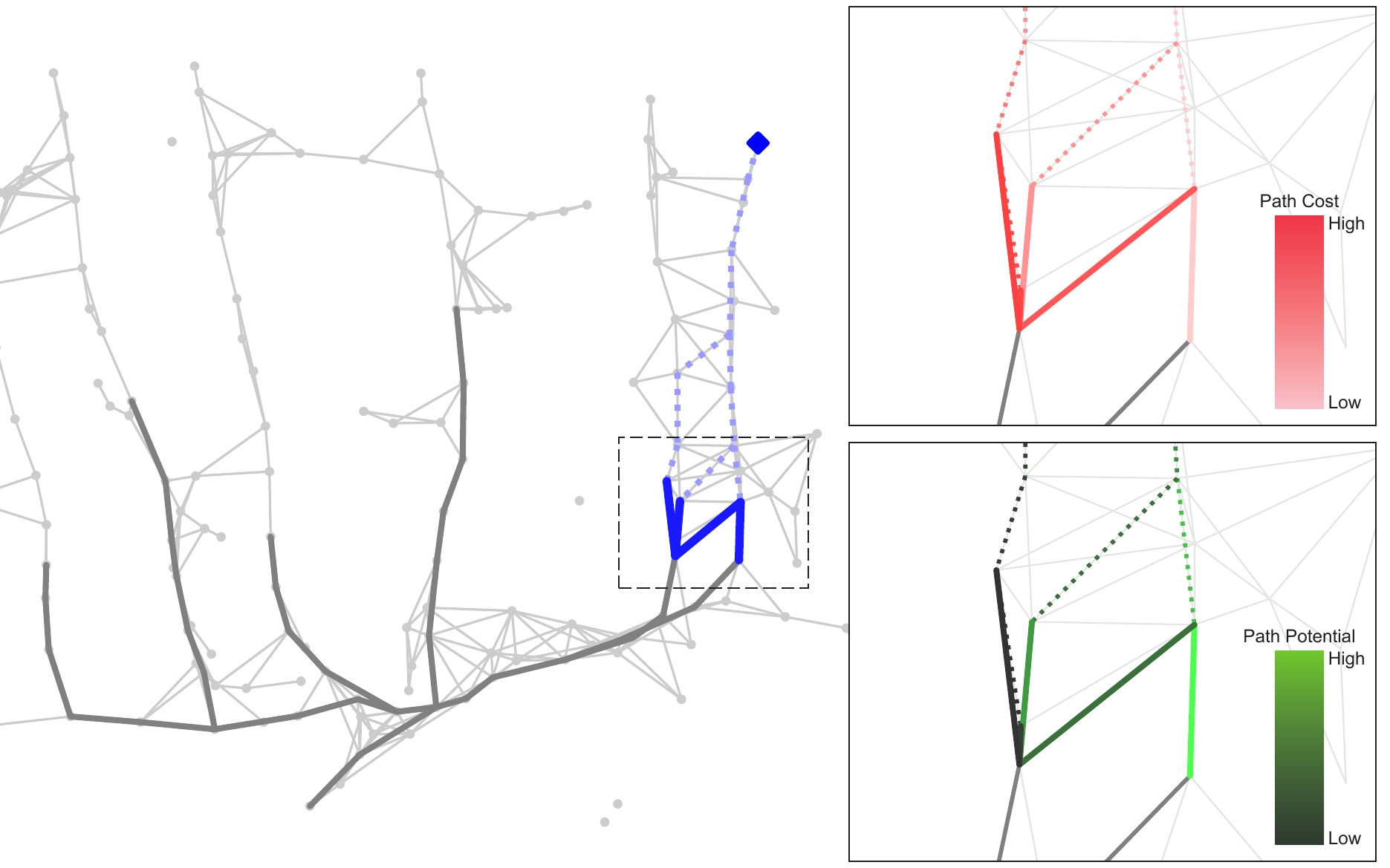}\caption{An example of how eligible edges to be added to the skeleton candidate are determined. The edges which can be added to the existing skeleton are shown in blue (Section~\ref{sec:edgeproposal}), with their corresponding tip path costs and potentials shown on the right (Section~\ref{sec:potential}). Note the correlation between low cost and high potential edge candidates.}\label{fig:costpotential}
    \vspace{-2ex}
\end{figure}

This section considers the core problem of taking an existing skeleton candidate $S$ and adding an edge to it, simultaneously assigning the edge a label. In particular, we would like to be able to \textit{rank} edge-label pair additions to understand which edge-label pairs are better than others, which we discuss in Section~\ref{sec:potential}. First, however, we must determine what edge-label pairings are valid additions to the existing candidate skeleton $S$. The key insight we utilize in this section is the assumption that, disregarding the side branches, the final skeleton will be the union of all the skeleton's base-to-tip paths. Then given a single tip superpoint $n_{tip}$ (which we assume is chosen randomly), our goal is to identify topologically-valid edge-label pairs which we can add to $S$ which are capable of growing towards $n_{tip}$, as shown on the left of Figure~\ref{fig:costpotential}. 

The key to identifying eligible edges is to use Dijkstra's algorithm to precompute path priors for \textit{all} edges in $(N, \overline{E})$ which lead to $n_{tip}$. For simplicity and efficiency, when utilizing Dijkstra's algorithm, we do not consider the labels applied to the edges, only the edge confidences and geometry. As Dijkstra's algorithm is a cost-minimization algorithm, we define a cost function which applies a cost for two factors: edges with low confidence and sharp angles. Letting $e_s$ and $e_p$ be two adjacent edges in $\overline{E}$, the cost between two edges defined as follows:

\begin{equation}
    \label{eq:costfunc}
    \funcvar{Cost}(e_s, e_p) = \Len(e_s)\left(1 - \Conf(e_s)\right) + \funcvar{TurnPenalty}(e_s, e_p, L_{\funcvar{None}})
\end{equation}

(The $L_{\funcvar{None}}$ is a label mapping that maps all edges to the label $\funcvar{None}$, indicating that the cost function does not consider the labelling of the edges and always applies the turn penalty between adjacent edges.)

Having defined the cost function, we let $Dijk(n_{tip}, e)$ denote the edges of the minimum cost path computed by Dijkstra's algorithm from $e$ to $n_{tip}$; this may be the empty set if no such path exists. The edge $e$ is then eligible to add to the skeleton candidate $S = (N, E, L)$ if the following conditions are met:

\begin{itemize}
    \item $Dijk(n_{tip}, e)$ is non-empty.
    \item The addition of the path prior $Dijk(n_{tip}, e)$ to $E$ does not create a topology violation. (Section~\ref{sec:term})
    \item There exists at least one possible label $l \in \{{\tt Trunk, Support, Leader}\}$ such that the addition of $(e, l)$ to $S$ does not create a label topology violation. (Section~\ref{sec:topological})
\end{itemize}

We denote the set of all eligible edge-label pairs to add to $S$ which connect to $n_{tip}$ as $\funcvar{Eligible}(S, n_{tip})$.

\subsubsection{Evaluating edge-label pair additions}
\label{sec:potential}

After determining all eligible edge-label pairs, our next step is to determine which ones represent the best additions. For a skeleton $S = (N, E, L)$, we use the notation $S \cup (e, l)$ to describe the new skeleton with $e$ added to $E$ and $L(e) = l$. While we could simply use the objective function $\funcvar{SkelScore}$ to score $S \cup (e, l)$, evaluating edge-label additions solely based off of the objective score ignores the global context from before that added edges should grow towards tips. In this section, we define a metric called $\funcvar{Potential}$ which, given a tip superpoint $n_{tip}$ and a topologically valid edge-label pair $(e, l)$, takes into account not just the score of the resulting skeleton $S \cup (e, l)$, but the ability of $(e, l)$ to be part of a viable path towards the tip $n_{tip}$.

The key to defining the $\funcvar{Potential}$ metric is to use the path priors computed by Dijkstra's algorithm in the previous section. While the path of lowest cost to the tip computed by Dijkstra's algorithm is not guaranteed to be the same as the path of largest reward, by encouraging Dijkstra's algorithm to pick paths with high confidence and low turn angles, our path priors should be highly correlated with the paths of highest reward, as shown on the right of Figure~\ref{fig:costpotential}. The  $\funcvar{Potential}[S, n_{tip}]$ function, parametrized by a skeleton candidate $S$ and a randomly chosen tip superpoint $n_{tip}$, takes in an valid edge-label pair $(e, l)$ which can be added to $S$ and sums the following values:

\begin{itemize}
    \item The objective value of the resulting skeleton, i.e. $\funcvar{SkelScore}(S \cup (e, l))$.
    \item The edge scores for each of the edges in the path prior $Dijk(n_{tip}, e)$.
\item The turn penalties for each adjacent edge pair in the path prior $Dijk(n_{tip}, e)$. Depending on the label $l$, we drop $2 - Order(l)$ of the largest turn penalties: $2$ if $l = {\tt Trunk}$, $1$ if $l = {\tt Support}$, and $0$ if $l = {\tt Leader}$. (The $Order$ function is defined in Section~\ref{sec:topological}.)
\end{itemize}

\begin{align}
\begin{split}
    \funcvar{Potential}[S, n_{tip}](e, l) &= \funcvar{SkelScore}\left(S \cup (e, l)\right) + \sum_{e \in Dijk(n_{tip}, e)} \funcvar{EdgeScore}(e) \\
    & - \underbrace{\sum_{e \in Dijk(n_{tip}, e)} \funcvar{TurnPenalty}(e, Desc(e), L_{\funcvar{None}})}_{\text{Drop highest $(2 - Order(l))$ values from summation}}
\end{split}
\end{align}

The justification for dropping the turn penalties is as follows. Recall that $\funcvar{TurnPenalty}$ (Equation~\ref{eq:turnpenalty}) returns 0 if two adjacent edges have different labels. Furthermore, when considering an edge-to-tip path, our assumption about branches being generally straight implies that the sharpest turn angles will occur at such label transitions. When considering a path that reaches the top of a leader, the number of such label transitions that can occur when assigning an edge with a label $l$ is given by $Order({\tt Leader}) - Order(l) = 2 - Order(l)$. In lieu of trying to consider exactly where the transitions should occur and scoring the path prior accordingly, we simply assume that each remaining label transition corresponds with the largest turn penalty remaining.

\subsubsection{Proposing and resampling candidate skeletons}
\label{sec:pop_sample}

We now move on to the process of taking a population of skeleton candidates $\{S_{k,i}\}_{k=1}^K$ at iteration $i$ and creating the new set of candidates $\{S_{k,i+1}\}_{k=1}^K$ at iteration $i+1$. First, we compute the set of \textit{all} possible next-generation skeleton candidates. We then assign a weight to each of these candidates which favors candidates associated with a high score (Section~\ref{sec:objfunc}) and a high potential to reach a tip of the tree (Section~\ref{sec:potential}). For simplicity, we drop the iteration value $i$ from the notation for the remainder of this section.

To compute all the unique possible next-generation skeleton candidates, for each current skeleton candidate $S_k$, we choose a random tip superpoint $n_{tip,k}$ and compute the eligible set of edge-label pairs $\funcvar{Eligible}(S_k, n_{tip,k})$. We then add all skeletons which can be created from one edge-label addition from the eligible set to $S_k$ to the next-generation candidate set. More formally, the full set of next-generation candidates is defined as:

\begin{equation}
    \bigcup_{k=1}^K \left\{S_k \cup (e, l): (e,l) \in \funcvar{Eligible}(S_k, n_{tip,k})\right\}
\end{equation}

Our goal is to define a function $\funcvar{Weight}$ to each of these candidates which favors candidates whose parents had a high $\funcvar{SkelScore}$ and which have a high $\funcvar{Potential}$ to reach a tip superpoint. The key idea is that we convert the raw values from $\funcvar{SkelScore}$ and $\funcvar{Potential}$ into weighted \textit{ranks} which we use to control the likelihood of a next-generation candidate being chosen. First, we define a ranking function denoted $Rank\left(\funcvar{Func},\, x_j,\, \{x_k\}\right)$, where $x_j$ is an element in the set $\{x_k\}$, and $\funcvar{Func}$ is a mapping $\{x_k\} \rightarrow \mathbb{R}$. $Rank$ then returns a value from $\frac{1}{|X|}$ to $1$ in increments of $\frac{1}{|X|}$ corresponding to the 1-indexed position of $\funcvar{Func}(x_j)$ in the sorted list $\{\funcvar{Func}(x_k)\}$, with tied values having their ranks averaged together. For instance, if $\funcvar{Func}$ is the identity function and $\{x_k\} = \{4, 5, 5, 3, 7\}$, the corresponding values from $Rank$ would be $\left\{\frac{2}{5}, \frac{7}{10}, \frac{7}{10}, \frac{1}{5}, 1\right\}$.

We use the $Rank$ function to create a resampling weight that favors next-generation skeleton candidates which were proposed by high-scoring parents and that have the best potential to reach one of the tips of the tree. Specifically, let $S_j$ be a skeleton in the current iteration's population $\{S_k\}_{k=1}^K$, and let $n_{tip,j}$ be a randomly-chosen tip superpoint. Let $(e, l)$ be a proposed edge-label addition from the eligible set $\funcvar{Eligible}(S_j, n_{tip,j})$. We define a function $\funcvar{Weight}$ associated with adding the edge-label pair $(e, l)$ to $S_j$, outputting a value in the range $(0, 1]$:

\begin{align}
\begin{split}
    \funcvar{Weight}(S_j, e, l) & = Rank\left(\funcvar{SkelScore},\, S_j,\, \{S_k\}_{k=1}^K\right) \\
    & \times Rank\left(\funcvar{Potential}[S_j, n_{tip,j}],\, (e, l),\, \funcvar{Eligible}(S_j, n_{tip,j})\right)
\end{split}
\end{align}

We use this $\funcvar{Weight}$ function to assign each next-generation candidate a numeric value. Since a next-generation candidate may be proposed by multiple current skeletons, we simply sum the corresponding weights together, thereby also favoring new candidates which are commonly proposed by existing candidates. Afterwards, we simply rescale all of the weights to sum to 1 and perform a weighted random sampling to create the new population of skeleton candidates. We also introduce a parameter $k_{MaxRep}$ which represents the maximum number of times any unique skeleton maybe represented in the population, which aids in encouraging population diversity at the start of the process. This process of proposing grown skeletons and resampling from them continues until all of the skeletons have been fully grown towards all of the tip nodes.

\subsubsection{Locating side branches}
\label{sec:sidebranches}

\begin{figure*}[bt!]
	\centering
\includegraphics[width = 14cm]{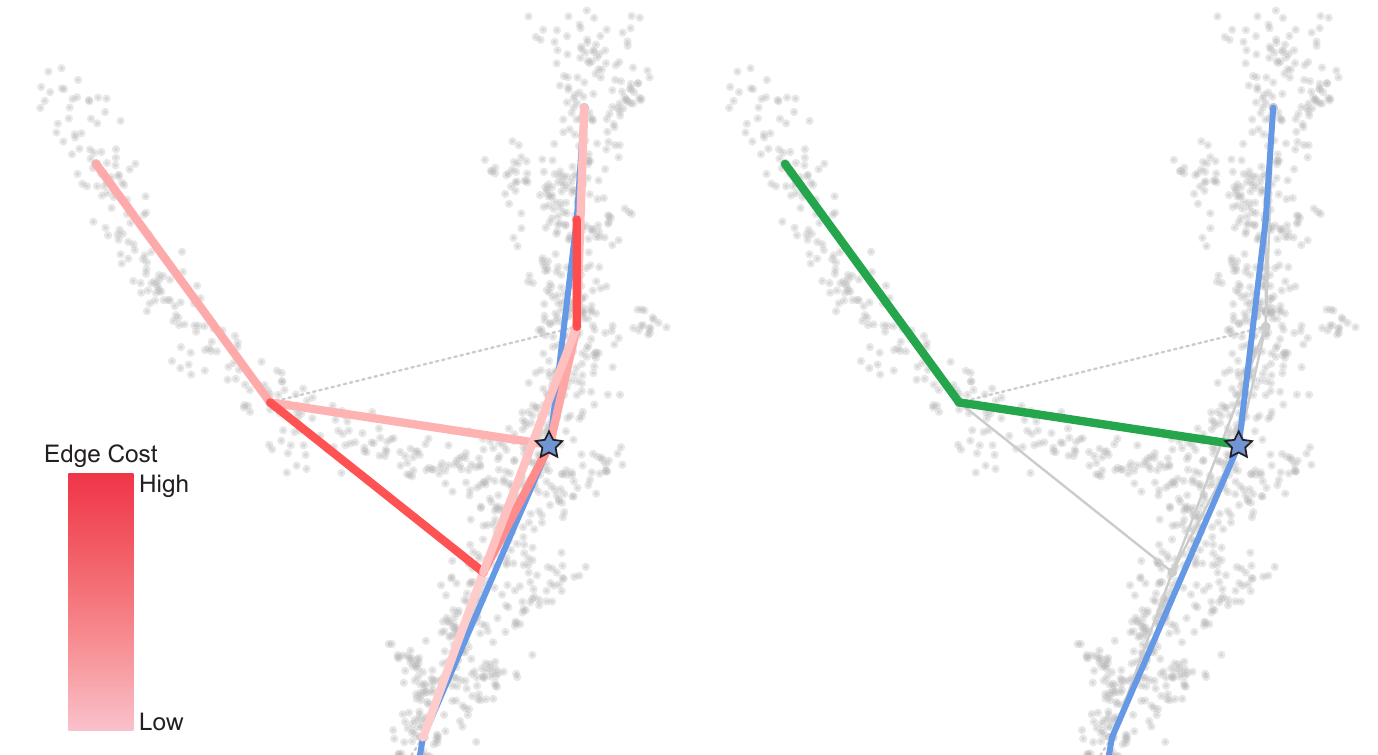}\caption{An illustration of finding side branches on the tree. First, an edge roughly orthogonal to a leader is identified (the blue star), and its corresponding connected component is identified, with edge costs shown in red. We then use a simple cost minimization search to identify the side branches.}\label{fig:sidebranches}
    \vspace{-2ex}
\end{figure*}

Once the iterative process has terminated, yielding an estimate of the optimal skeleton consisting of the trunk, support, and leaders, we can then run a simple search process to find the side branches as illustrated in Figure \ref{fig:sidebranches}. We define a side branch to be a branch which comes off of a leader branch at a right angle from the leader, plus or minus a tolerance of 45 degrees. We find all side branches through the following process:

\begin{enumerate}
    \item Remove all edges from the graph below the confidence threshold $\alpha_{conf}$.
    \item Identify all edges which branch off from a leader at an angle between 45 and 135 degrees, as well as their respective connected components.
    \item For each connected component, run Dijkstra's algorithm to find the path with the lowest total $\funcvar{Cost}$ within the connected component, subject to the constraint that the turn angle of any two adjacent edges cannot be greater than $\pi/2$ radians. The search terminates when the branch is grown out such that no further superpoint neighbors can be connected.
\end{enumerate}

\begin{table}[p]
\begin{tabular}{|p{0.16\linewidth}p{0.66\linewidth}p{0.08\linewidth}|}
\hline
Parameter & Description & Value\\ \hline
\multicolumn{3}{|c|}{\textbf{Initialization parameters}} \\
$r_{super}$ & Sphere radius used for superpoint construction. & $0.10$ \\ 
$\alpha_{tip}$ & Threshold indicating $z$ range of superpoints to be checked for tip nodes. & $0.60$ \\
$\alpha_{conf}$ & Minimum confidence parameter. Edges with confidence below this will have negative reward and may be discarded in the growing phase. & $0.40$ \\ \hline
\multicolumn{3}{|c|}{\textbf{Turn angle penalty polynomial parameters}} \\
$\theta_{TurnMin}$ & Minimum turn angle for edges with the same label for penalty to be applied. & $\pi/4$ \\
$c_{Turn}$ & Coefficient on turn angle penalty. & $0.5$ \\
$p_{Turn}$ & Power on turn angle penalty. & $2$ \\ \hline
\multicolumn{3}{|c|}{\textbf{Growth angle penalty polynomial parameters.}} \\
$\theta_{GrowMin}$ & Minimum growth angle deviation for penalty to be applied. & $\pi/4$ \\
$c_{Grow}$ & Coefficient on growth angle penalty. & $0.3$ \\
$p_{Grow}$ & Power on growth angle penalty. & $1.0$ \\ \hline
\multicolumn{3}{|c|}{\textbf{Population resampling parameters}} \\
$K$ & Population size. & $500$ \\
$k_{MaxRep}$ & Maximum allowable duplicates of skeleton candidates. & $3$ \\
\hline
\end{tabular}
\caption{Parameter table.}
\label{table:params}
\end{table}

\begin{table}[p]
\begin{tabular}{|p{0.23\linewidth}p{0.69\linewidth}|}
\hline
Notation & Description\\ \hline
\multicolumn{2}{|c|}{\textbf{Skeleton elements}} \\
$P=\{p_i\}$ & A set $P$ of points $p_i \in \mathbb{R}^3$ representing the point cloud\\
$N = \{n_i\}$  & A set $N$ of superpoint node points $n_i \in \mathbb{R}^3$ \\
$E = \{e_j\}$ & A set $E$ of directed edges $e_j = (n_{j_1}, n_{j_2})$, where $n_{j_1}, n_{j_2} \in N, n_{j_1} \neq n_{j_2}$ \\
$L(E)$ & A mapping of edges in $E$ to labels in the set $\{\funcvar{Trunk}, \funcvar{ Support}, \funcvar{ Leader}, \funcvar{ SideBranch}, \funcvar{ None}\}$\\
$S = (N, E, L)$         & A labelled skeleton composed of its superpoints, edges, and labels for each edge \\ 
$n_B$ & The superpoint in $N$ that represents where the base of the trunk of the tree meets the ground. \\ \hline
\multicolumn{2}{|c|}{\textbf{Edge characteristics}} \\
$\funcvar{Len}(e)$, $e = (n_{i_1}, n_{i_2})$  & $\lVert n_{i_1} - n_{i_2} \rVert$ \\
$\funcvar{GrowAngle}(e) \in [0, \pi/2]$ & Angle formed between $x$-axis and edge projected into the $xz$ (ground) plane\\
$\angle(e_s, e_p)$ & Turn angle between an edge $e_s$ and its predecessor $e_p$. ($e_p$ is denoted $Pred(e_s)$ when $E$ is fixed.) \\
$\Conf(e) \in [0,1]$ & Neural network score, confidence of edge \\ \hline
\multicolumn{2}{|c|}{\textbf{Objective function}} \\
$\funcvar{SkelScore}(S)$ & $\sum_{e \in E} \funcvar{Reward}(e, L)$ \\
$\funcvar{EdgeScore}(e)$ & $\Len(e) \left(1 - \frac{1 - \Conf(e)}{1 - \alpha_{conf}}\right)$ \\
$\funcvar{TurnPenalty}(e_s, e_p, L)$ & Penalty induced by the addition of $e_s$ to the skeleton due to the turn angle formed with its predecessor edge $e_p$\\
$\funcvar{GrowPenalty}(e, L)$ & Penalty induced by the addition of $e$ to the skeleton due to the growth angle of the edge \\
$\funcvar{Reward}(e, L)$ & $\funcvar{EdgeScore}(e) - \funcvar{TurnPenalty}(e,\, Pred(e),\, l) - \funcvar{GrowPenalty}(e, L)$ \\ \hline
\multicolumn{2}{|c|}{\textbf{Auxiliary metrics}} \\
$\funcvar{Cost}(e_s, e_p)$ & $ \Len(e_s)\left(1 - \Conf(e_s)\right) + \funcvar{TurnPenalty}(e_s, e_p, L_{\funcvar{None}})$\\ 
$\funcvar{Potential}[S, n_{tip}](e, l)$ & Approximate reward associated with new skeleton $S \cup (e, l)$ plus a path prior from $e$ to $n_{tip}$ \\
$\funcvar{Weight}(S_j, e, l) \in (0, 1]$ & Resampling weight which combines the relative $\funcvar{SkelScore}$ of $S_j$ with the relative $\funcvar{Potential}$ of $(e, l)$ to reach a tip node \\ \hline
\end{tabular}
\caption{Important notation and equations used throughout the paper.}
\label{table:notation}
\end{table}


\section{Experimental setup}
\label{sec:experiment}

\subsection{Data collection}
\label{sec:datacollection}

\begin{figure}[h]
	\centering
\includegraphics[width = 90mm]{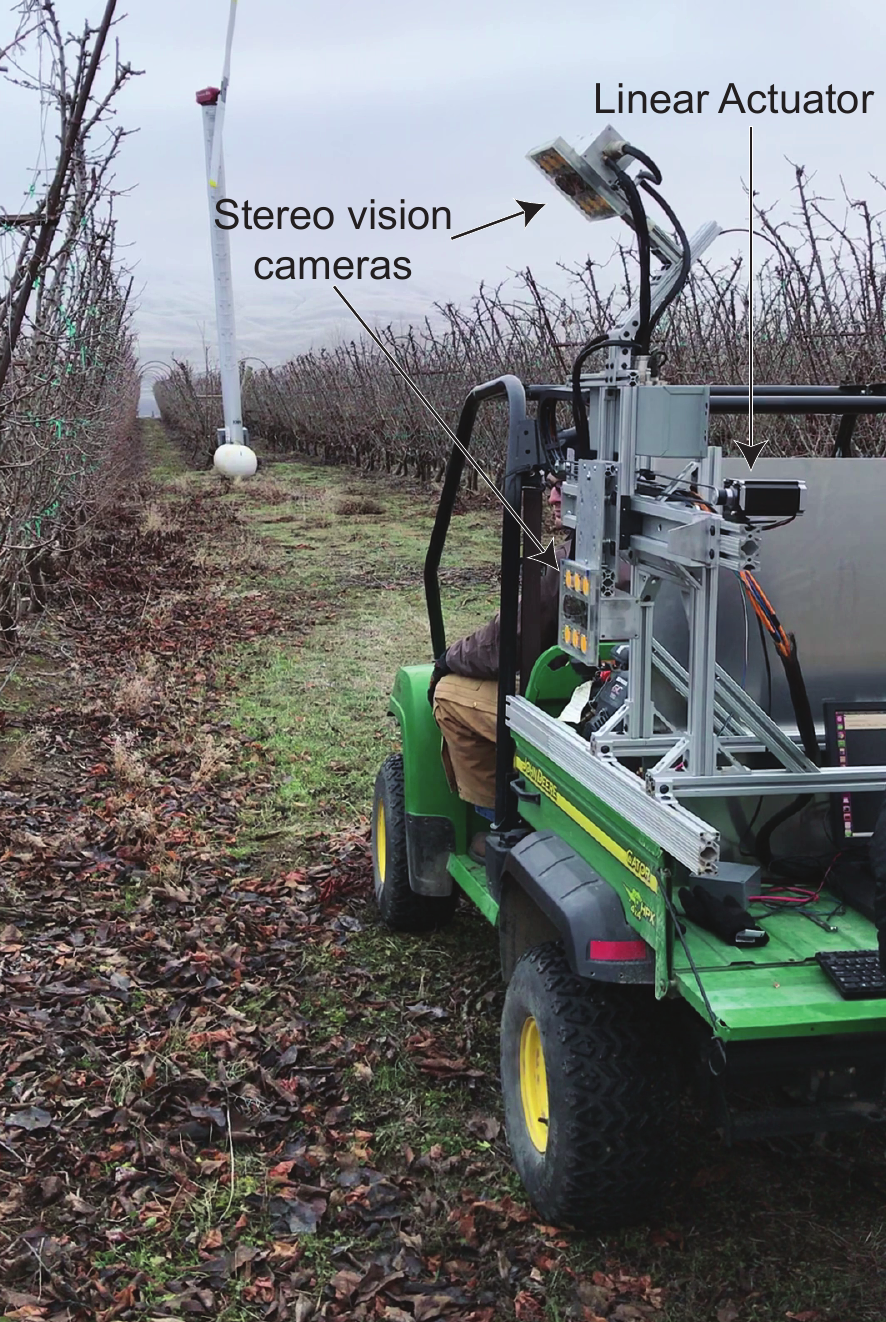}\caption{The multi-camera setup which was used in the field to produce point clouds.}\label{fig:camerasetup}
    \vspace{-2ex}
\end{figure}

The custom sensor array that we used to generate point clouds is shown in Figure \ref{fig:camerasetup}. The system includes a pair of stereo vision sensors arranged vertically with the top sensor angled down. A linear actuator consisting of a stepper motor and belt-driven carriage on a slide moves the sensor array to 3 preset positions. When recording a tree, at each of the three locations, the system takes an image; it then registers all of the images together to form the point cloud.

This system was mounted onto an electric utility vehicle that we took to a UFO sweet cherry orchard (Prosser, WA, USA) in December 2019. Our process consisted of stopping the vehicle in front of each tree in a row and allowing the scanning system to acquire images. We did this for a single row of trees, covering both sides of the row and ultimately ending up with 82 complete point clouds. The dataset is available \href{https://drive.google.com/drive/folders/1V2XNrYTp715YA0iOg8Ewn-oLq1Bha6m_}{at this link}. (Full URL: \url{https://drive.google.com/drive/folders/1V2XNrYTp715YA0iOg8Ewn-oLq1Bha6m_})

\subsection{Methodology}

From the 82 point clouds that we collected, we manually sorted through them and selected 43 point clouds without significant defects. Usually, each point cloud contained points from neighboring trees, including trees on the other side of the trellis, as well as the ground; therefore, we performed manual filtering to remove any points associated with the ground and neighboring tree.

From these filtered point clouds, 14 were selected for tuning our algorithm parameters, as well as training the edge validity neural network. After finalizing the parameters, we ran the algorithm on the remaining \numtrees point clouds, each of which was randomly downsampled to 50000 points three times, for a total of 87 evaluations of the algorithm.

Normally, evaluation of the algorithm would be done against a ground truth skeleton. However, a ground truth skeleton is difficult to define, as each evaluation will have a different superpoint graph. Furthermore, the point clouds themselves are sufficiently messy that even a human may have difficulty evaluating whether a section of the point cloud corresponds to a real branch or not. Therefore, to evaluate the algorithm, we obtained the help of two researchers familiar with the UFO tree structure but otherwise unaware of the inner workings of the algorithm. For each of the \numtrees trees, we presented them with the skeletonization algorithm outputs for the 3 trials and asked them to mark areas on the skeletons which were notably incorrect. We then applied the best-fitting corrections to the graph to create the corrected version of the skeleton.

To measure the difference between the skeleton produced by the algorithm $S = (N, E, L)$ and the corrected version $S^* = (N, E^*, L^*)$, we use the notion of an \textit{edit distance} to compute how many modifications need to be made to the original skeleton to result in the final skeleton. A single edit action includes removing an edge-label pair, adding an edge-label pair, or modifying the label on the existing skeleton. Therefore, the \textit{global} edit distance between $S$ and $S^*$ is the number of edges in the symmetric difference of $E$ and $E^*$ plus the number of edges in $E \cap E^*$ for which $L(e) \neq L^*(e)$. To make global edit distances comparable between skeletons of sizes, we normalize the edit distances by the number of edges in the corrected skeleton. 

We also examine the similarity of the skeletons on a per-label basis to see if the skeletonization algorithm has difficulties with some labels more than others. First, we subset each skeleton down to only the edges whose corresponding labels are $l$. We then compute the corresponding edit distance, which is equal to the size of the symmetric difference of the subsetted skeleton edges. Normalizing these edit distances is a bit more tricky than in the global case, as for some trees the number of ground truth labelled edges for certain labels can be very low (namely trunks and side branches), leading to inflated edit ratios. Therefore, we normalize the per-label edit distances by the number of \textit{contiguous} label segments in the ground truth tree, times the \textit{average} number of edges per contiguous label segment across all ground truth trees. For instance, if a ground truth tree has 8 distinct leader segments, and the average leader segment across all trees has 10 edges, then the denominator would be 80. All trunks are assumed to have just one contiguous trunk and support segment, whereas the number of leaders and side branches may vary. 








\newpage
\section{Experimental results}
\label{sec:results}


\begin{figure}[bt]
	\centering
\includegraphics[width = \columnwidth]{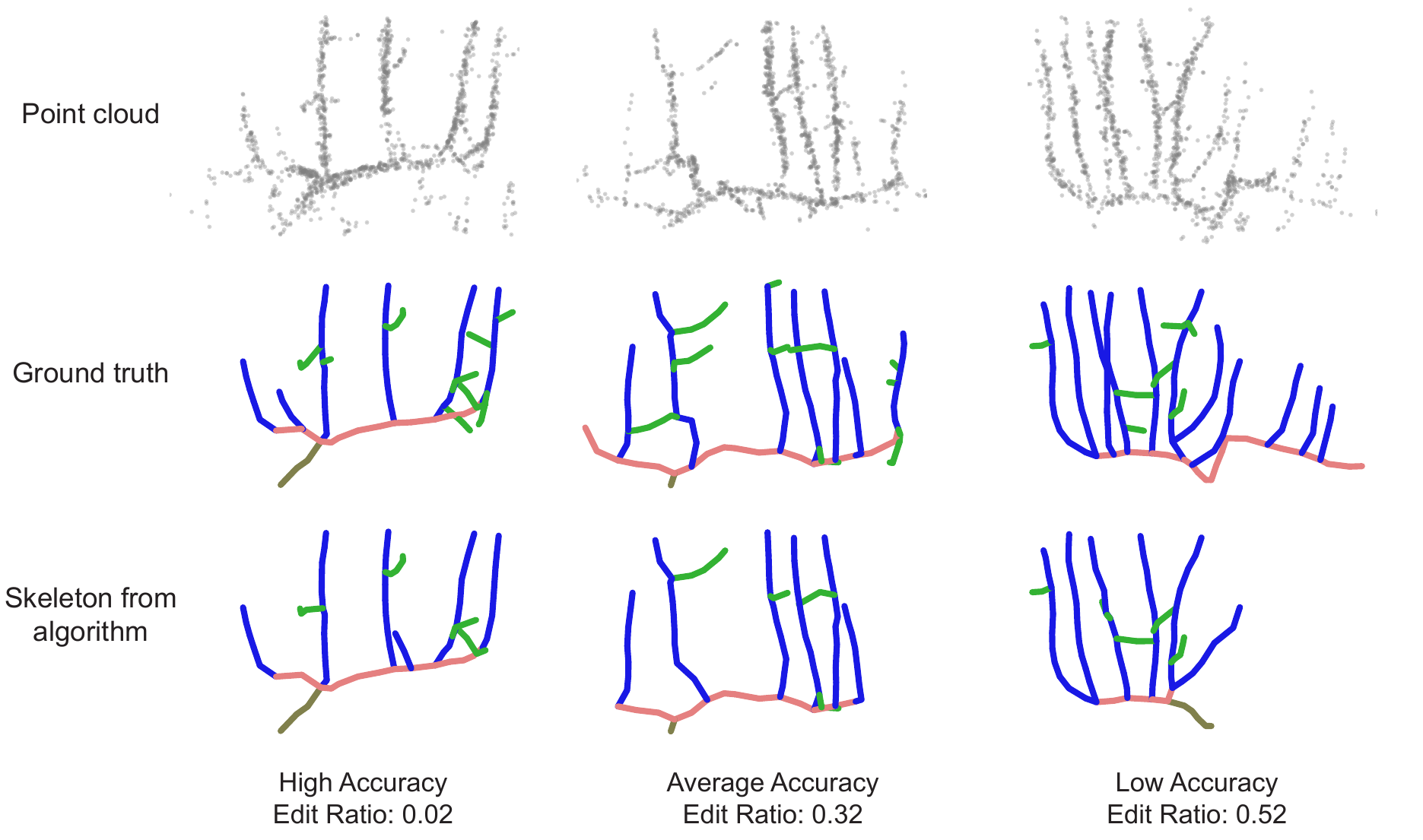}\caption{Examples of various skeletons which were generated by our algorithm (bottom row), along with the ground truth skeletons (middle row) and the point clouds from which they were generated (top row).}\label{fig:skeletons}
\end{figure}

\begin{figure}[p]
	\centering
\includegraphics[width = 0.75\columnwidth]{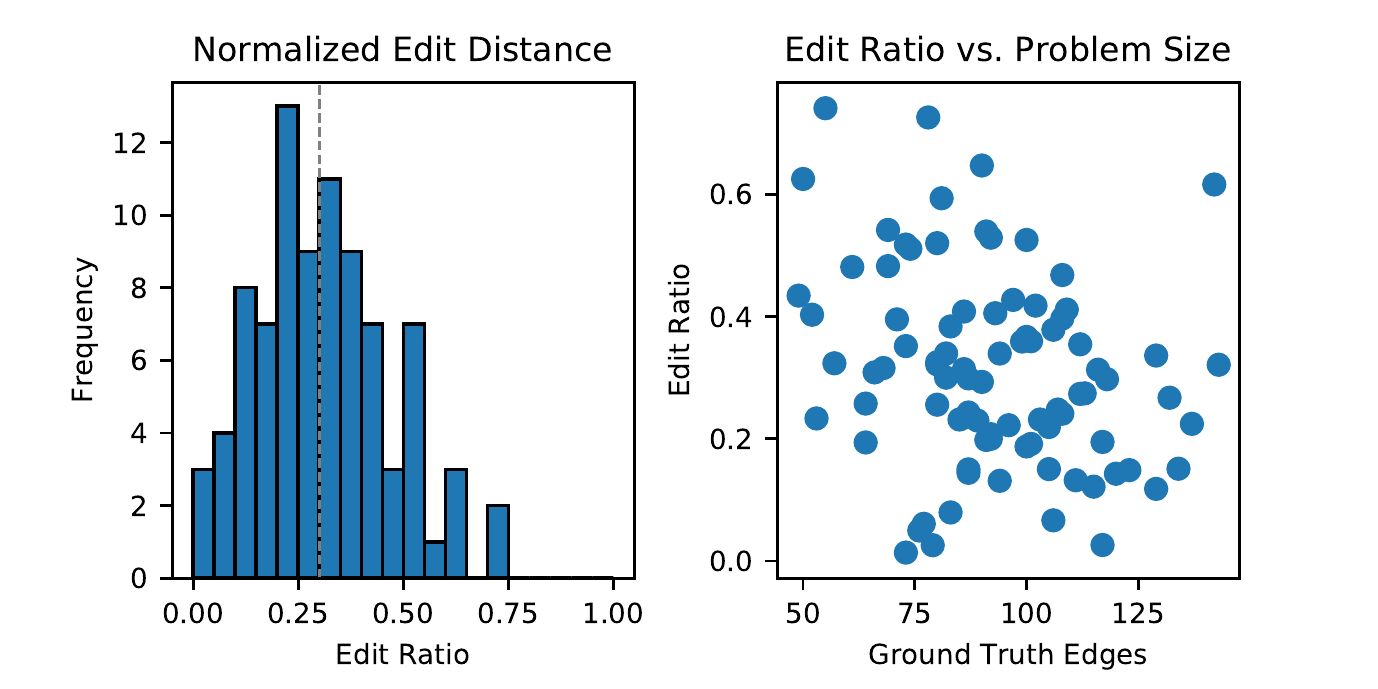}\caption{Left: The distribution of achieved global edit ratios for our skeletonization algorithm across 87 trials, with the median shown as a dotted line. Right: Achieved edit ratios versus the number of edges contained in the ground truth skeleton. }\label{fig:editglobal}
\end{figure}

Figure~\ref{fig:skeletons} shows some of the skeletons which were produced by our algorithm along with the corrected skeletons. Collectively, we were able to achieve a median edit ratio of 30\%, i.e. an accuracy of 70\%, with the best and worst edit ratios achieved being 1.4\% and 74\% respectively. Figure~\ref{fig:editglobal} shows the distribution of global edit ratios across all 87 trials. In general, the edit ratio distribution is skewed towards high edit ratios, i.e. while most skeletons required fairly small numbers of edits, there were also instances where significant portions of the generated skeleton needed to be fixed. We also include a scatter plot illustrating the relationship between the achieved edit ratio versus the number of edges in the ground truth skeleton. There is actually a slightly negative correlation between the two stats, suggesting that the errors that the algorithm makes are induced by factors other than the size of the problem. There may also be a bias effect in which smaller point clouds are scrutinized more thoroughly than larger trees with many edges.

\begin{figure}[p]
	\centering
\includegraphics[width = 0.75\columnwidth]{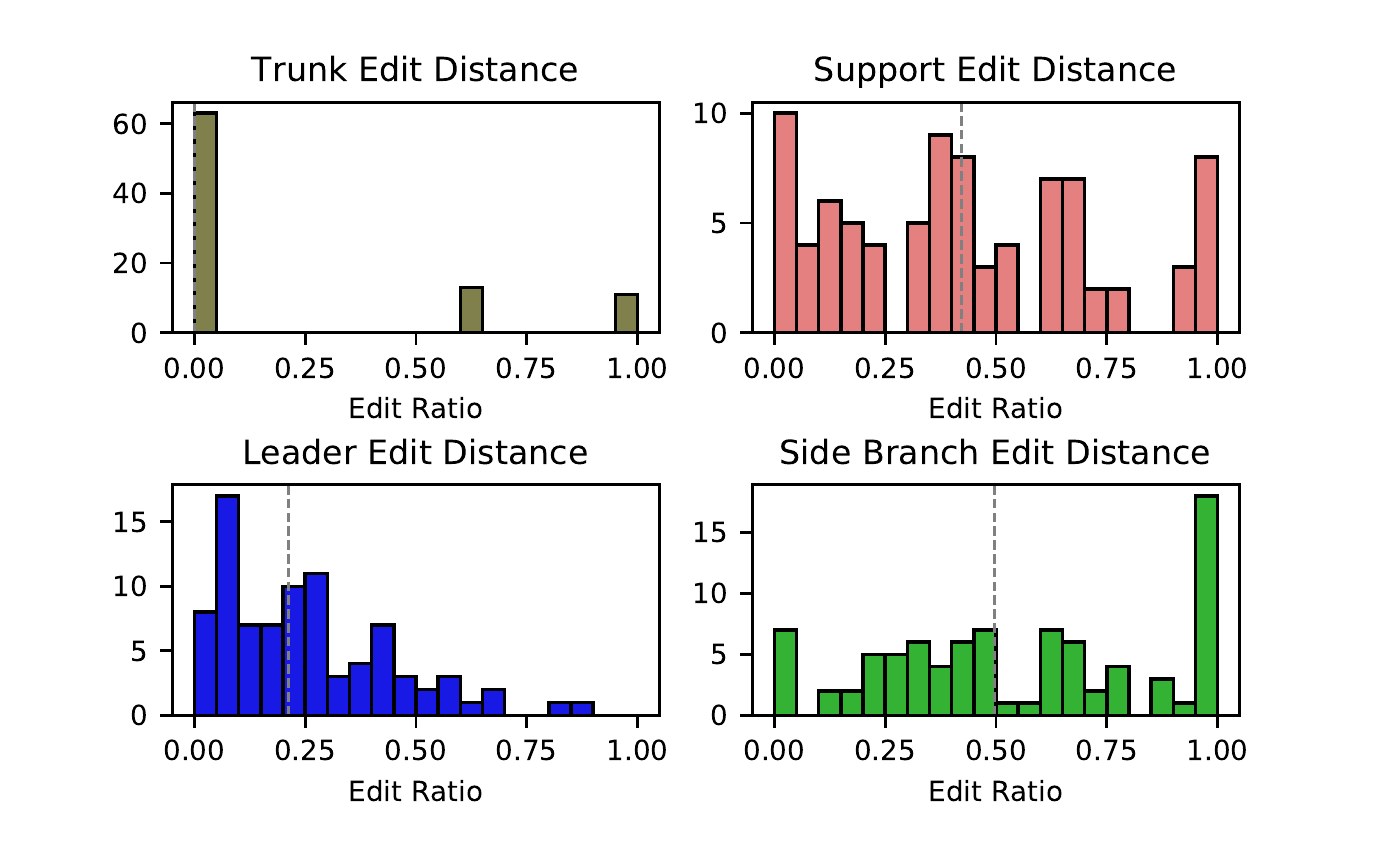}\caption{The distributions of normalized edit ratios on a per-label basis, for trunks, supports, leaders, and side branches. Medians are shown as a dotted line. All edit ratios greater than 1.00 have been grouped into the final bucket.}\label{fig:editlabel}
\end{figure}

\begin{figure}[p]
\centering
\includegraphics[width = \columnwidth]{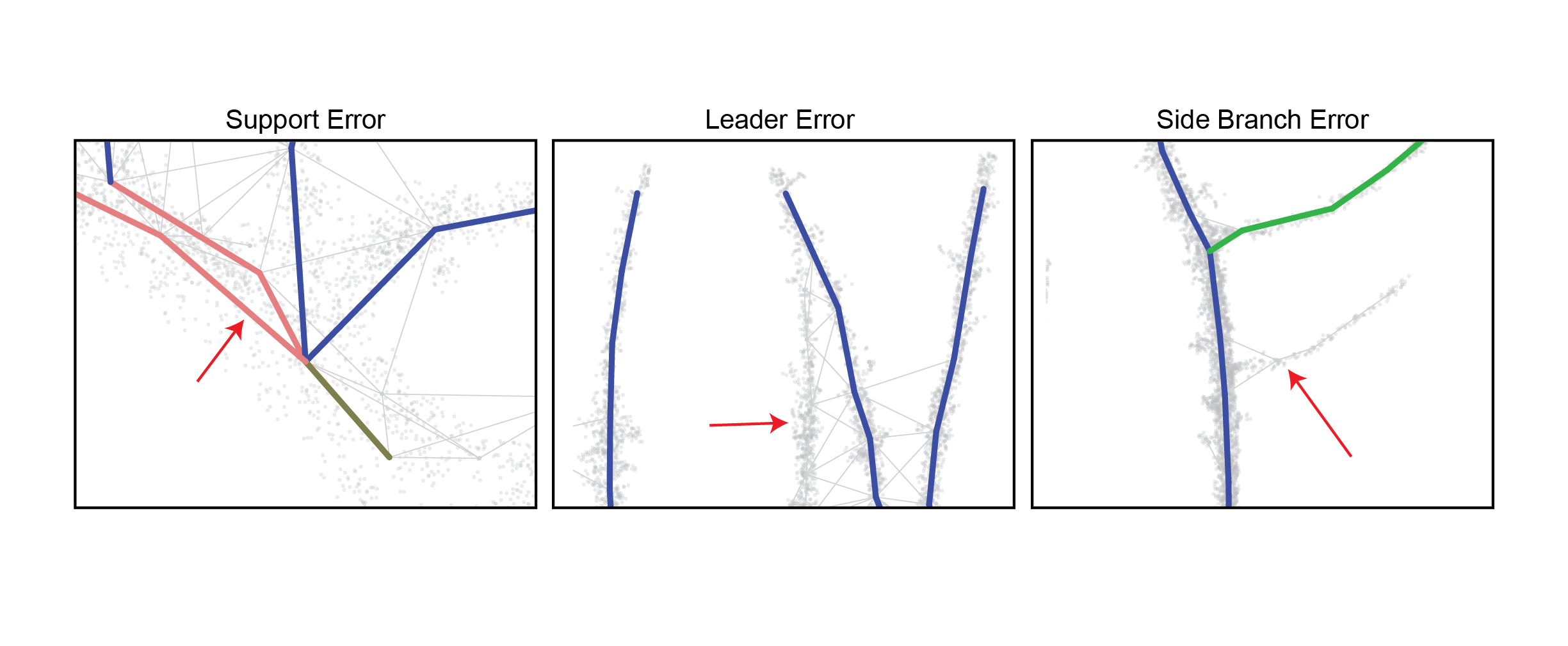}
\caption{Some examples of mistakes made by the skeletonization algorithm. Left: Both halves of the support (shown in red) were grown out in the same direction, resulting in one side of the tree being missed. Middle: A leader was missed due to two leader tips being close together. Right: A side branch was not detected due to the absence of a superpoint at the junction of the leader and side branch.}
\label{fig:errordiagram} 
\end{figure}

Next, we examine the normalized edit ratios for each of the four label types (trunk, support, leader, and side branch) shown in Figure~\ref{fig:editlabel}. For ease of comparability, all edit ratios greater than 1.0 have been grouped into the final bucket in the histogram. In general, the distribution of edit ratios varies quite a bit by label type:

\begin{itemize}
    \item Trunk - The median edit ratio was 0.00, i.e. typically no edits were necessary to the trunk. This is in itself not too surprising, as the average trunk contained just 1.6 edges. Larger edits were usually caused by the trunk overextending and splitting off into a support at an incorrect location.
    \item Support - The median edit ratio was 0.42. The formation of the support was a critical source of failures in the tree formation, as if the support stopped too early or grew off in the wrong direction, it would prevent entire leaders from being added to the skeleton. An example of an incorrect support formation is shown in the leftmost image in Figure~\ref{fig:errordiagram}, in which both halves of the support extended out in the same direction from the trunk, causing the other half of the tree to be missed.
    \item Leaders - The median edit ratio was 0.21. In general, leader detection was fairly accurate. Most required edits to leaders involved adding in leaders which were undetected because the support did not extend far enough. There were also occasional instances of pairs of leaders whose tips were sufficiently close together that only one of the two tips was properly detected, resulting in one of the leaders not being detected. An example is shown in the middle image of Figure~\ref{fig:errordiagram}.
    \item Side branches - The median edit ratio was 0.50. Side branch detection was the weakest of all of the four different labels. One of the main issues with detecting side branches is that in order for a side branch to be properly detected, a superpoint must be centered at the junction where the side branch meets the leader; otherwise, the edge will cut across diagonally from the leader to the side branch and look like a false connection, such as shown in the right image of Figure~\ref{fig:errordiagram}. Our superpoint generation process currently doesn't take this into account. Furthermore, some large edits were caused by the fact that if part of a structural element such as a support was missed in the skeleton, it would sometimes be misidentified as a side branch in the post-processing step, resulting in a large number of required edge removals. 
\end{itemize}

\begin{figure}[p]
	\centering
\includegraphics[width = 0.50\columnwidth]{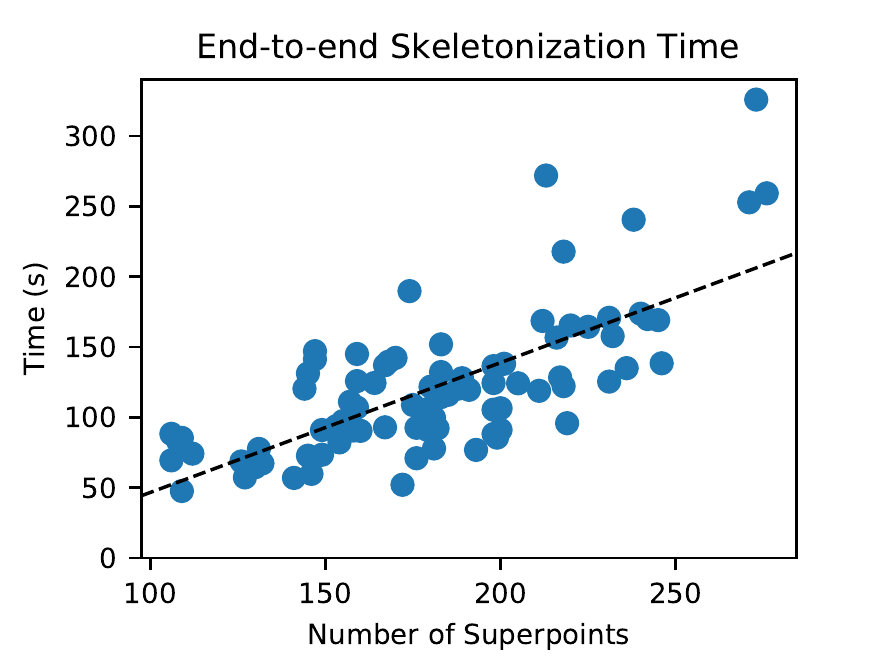}\caption{Runtime of the skeletonization algorithm with respect to the number of points in the underlying superpoint graph.}\label{fig:runtimes}
\end{figure}

Finally, we examine the relationship between the runtime of the skeletonization algorithm and the size of the problem, i.e. the number of superpoints contained in the superpoint graph. A scatter plot of the this relationship is shown in Figure~\ref{fig:runtimes}. The average split in the runtime between the pre-processing step and the search step was 5.2\% and 94.8\% respectively, with the mean total runtime across all trials being 120.8 seconds.

In general, the runtime trend appears to be roughly linear in the number of superpoints in the graph. This is not surprising given our assumption that the superpoint radius $r_{super}$ is chosen to be sufficiently large to fully encapsulate most branches, implying that the number of edges in the superpoint graph should scale linearly with respect to the number of superpoints in the graph generated from the tree point cloud. The main part of the algorithm which complicates this is the usage of Dijkstra's algorithm which, assuming the number of edges is proportional to the number of superpoints, has an $O(|E|\log|E|)$ complexity. However, since each of the Dijkstra's maps for each tip candidate are computed just once at the beginning of the optimization process, the time contributed from utilizing Dijkstra's algorithm is significantly outweighed by the search process itself for any reasonable size of $|E|$.

\newpage
\section{Discussion}
\label{sec:discussion}

While our skeletonization algorithm is not yet robust, our results demonstrate that in general, the algorithm is capable of extracting the majority of the semantic structure of UFO cherry trees. Many of the mistakes are sufficiently minor that they can be quickly corrected by a human. Furthermore, we can use the labelled results for further refinement of the optimization process if we wish to apply learning-based decision techniques in the future. Regarding the algorithm's usefulness for dormant pruning, we believe that minor improvements to the skeletonization process, primarily more informed placement of superpoints, should lead to better side branch detection. Furthermore, a post-processing step in which we refit the skeleton to the point cloud should allow us to determine the locations of the actual cut points on the side branches with sufficient accuracy for us to integrate our previous motion planning and control architecture. It would also be relatively simple to incorporate estimates of branch width into the CNN. Not only would this be useful for identifying the trunk base location automatically, but since tree branches generally become thinner the further up the tree they are, branch width heuristics could be incorporated into the objective to inform the construction of the skeleton.

One major possibility to improve our algorithm is to have the optimization process utilize probabilistic priors on the edge labels in order to steer the population towards more optimal skeleton candidates. We did attempt to compute label priors using the CNN as described in the Appendix, but they were not robust enough to be useful. Furthermore, we should be able to update these priors based off of the topology of the existing skeleton, as past assignments should inform the assignment of subsequent edges. For instance, the support mistake shown in the left of Figure~\ref{fig:errordiagram}, in which both of the supports grow out in the same direction, could have been avoided if the optimization understood that it is unlikely for a second support branch to grow out in the same direction as an existing one. Our current framework does not attempt to hardcode these sorts of assumptions into the objective function, instead assuming that population diversity will keep skeleton candidates with the correct support split in the population. However, it appears that in general, there is no guarantee that this will be the case, especially when it may take many iterations for the objective function to discern that a previous assignment was suboptimal, by which point all correct assignments may have fallen out of the population. 

Finally, it is useful to consider the applicability of our algorithm to other types of trees, especially since past skeletonization algorithms have focused on moving away from heuristics and assumptions about structure. While some of the topological rules and penalties we defined are specific to UFO cherry trees, some of them also hold across many types of trees, such as our assumptions on branch turn angles being small. Furthermore, if the goal is labelled skeletonization, each type of tree will require some set of rules specific to that kind of tree which must be incorporated into the skeletonization process. So long as the structure of the tree can be recovered by growing the tree towards a set of tip superpoints, our framework should apply to any type of tree so long as the appropriate penalties are incorporated into the objective function. However, an even more optimal solution would be for our system to \textit{learn} the topology and geometry-based rules based off of priors instead of having to manually define constraints and penalties. 


\section{Conclusions}

In this paper, we introduced an algorithm which takes in a 3D point cloud and outputs a labelled skeleton that not only accurately represents the topology of the point cloud, but also provides semantic information on the different parts of the tree. It does so via a population-based search method that builds up candidate skeletons one edge-label pair at a time, incorporating both local and global context to determine which candidates are the best. The optimization function makes use of a robust correctness metric output from a CNN to ensure that only valid edges are added to the final skeleton. In our experiments, we were able to achieve \accperc accuracy with our skeletonization system, demonstrating that such a skeletonization approach is viable and could potentially be used directly in a pruning application. We also make our dataset of UFO cherry tree point clouds public, accessible \href{https://drive.google.com/drive/folders/1V2XNrYTp715YA0iOg8Ewn-oLq1Bha6m_}{at this link}.

In the immediate future, our plans are to use this framework during outdoor field trials of an end-to-end, robotic tree pruning system by linking the pipeline discussed in this paper with our previously developed motion planning and control architecture~\citep{you2020}. Long-term research into improving the skeletonization process would involve increasing generalizability of the algorithm via learning methods, as well as improving the performance of the search algorithm via context-aware priors on edge likeliness and labels, which should improve the algorithm's ability to quickly discover optimal solutions.



%

\balance
\bibliography{references}
\bibliographystyle{plainnat}



\newpage

\appendix

\section{Neural network details}
\label{sec:nn}
As noted in Section \ref{sec:edgeeval}, we use a CNN to evaluate the validity of the edges in a superpoint graph by taking the points associated with each edge, projecting them into a 2D fitting plane, and binning them into a 32 x 16 histogram to form a greyscale image. This section covers the architecture of our network, as well as our training methods.

\subsection{Architecture}

\begin{figure}[bt]
	\centering
\includegraphics[width = 0.9\columnwidth]{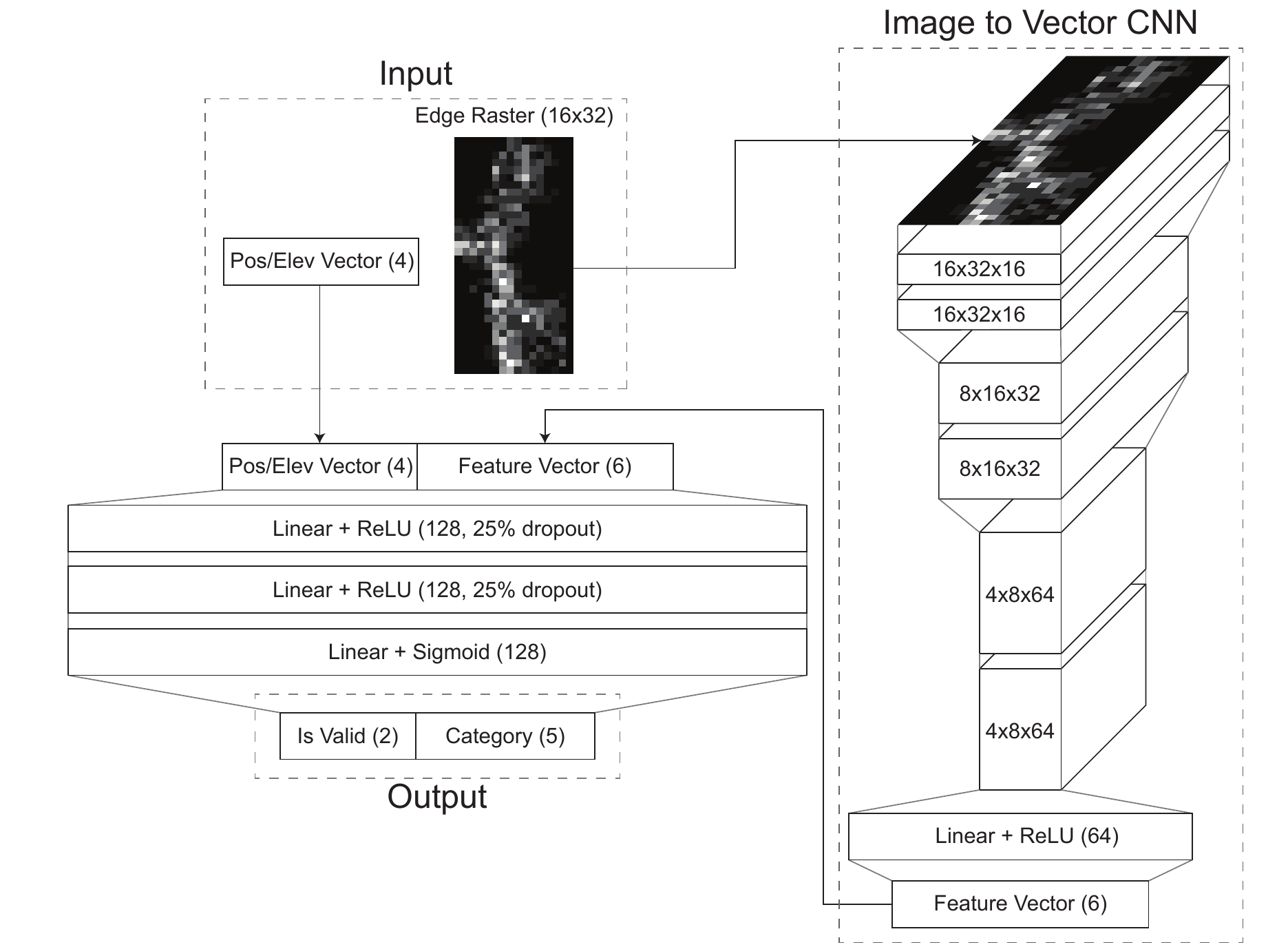}\caption{The architecture of our neural network which takes in two inputs, a 16 x 32 single-channel image and a description vector containing the branch center's location in $\mathbb{R}^3$ and elevation. It returns two prediction vectors: One for whether or not the corresponding edge is valid, and one predicting whether the edge is a trunk, support, leader, side branch, or something else.}\label{fig:cnnarchitecture}
    \vspace{-2ex}
\end{figure}

The main architecture of our network is shown in Figure \ref{fig:cnnarchitecture}. The first part of the network is a CNN which converts the 16 x 32 raster image into a 6-vector. All of the convolutional and feedforward layers use a rectified linear unit (ReLU) as the activation function; the convolutional layers have a filter size of 3. First, the image is processed by two convolutional layers with 16 channels each, after which we perform a batch normalization and a max pooling operation of size 2. We then do this again with two 32-channel convolutional layers. Finally, we have 2 more 64-channel convolutional layers. After these layers, instead of another batch normalization and max pool, we feed the feature maps into a feedforward network with one hidden layer with 64 nodes. The result is a 6-vector which serves as a compressed representation of the input image.

Once we compute the feature vector from the raster image, we append it with the 4-element description vector to obtain a 10-vector, which is then fed through a feed forward neural network with three hidden layers, each of which has 128 nodes in it. For the first two layers, we also apply a 25\% dropout operation. After passing the final layer through a sigmoid function, the final output is a 7-vector consisting of prediction values in $[0, 1]$. Two of these values represent confidences for the edge validity, and the remaining five represent the confidence values for each of the five edge categories assuming the edge is valid.

\subsection{Network Training}

In order to train the network, we utilized a training set of 14 point cloud trees. For each tree, we generated several different superpoint graphs and manually labelled the validity and class of the displayed segment. We augmented our data by saving the location of the endpoints of each edge, allowing us to create new rasterized images associated with the corresponding edge simply by resampling the point cloud. We did this for point clouds between 20000 and 100000 points in order to capture a variety of point densities. We also resampled the the input data so that all of the classes had a roughly uniform distribution. In total, we generated 15000 data points to use for training and validating the network, which we split 70\% for training, 20\% for validation, and 10\% for testing. For training the network, we use an Adam optimizer~\citep{kingma2014adam} with cross-entropy loss.

When computing the loss, it is important to note we have two different losses: One for the edge validity and one for the edge category. However, we note that that edges which are invalid do not have a category associated with them; therefore, when computing the category loss, we only backpropagate the loss associated with edges which actually represent valid connections. After each epoch of training, we compute the validity and category losses on the validation set and weight the losses by $2/3$ and $1/3$ respectively to create a combined loss metric. We then save the model if its combined loss is lower than that of the previously saved model.


\subsection{Training Results}

\label{sec:atraining}

\begin{table}[]
\centering
\begin{tabular}{|l|ll|}
\hline
& Validation & Testing \\ \hline
Edge Validity & 91.0\% & 89.8\% \\
Edge Category & 63.5\% & 59.1\% \\ \hline
\end{tabular}
\caption{Overall accuracy results for the neural network.}
\label{table:trainingresultsoverall}
\end{table}

\begin{table}[]
\centering
\begin{tabular}{|l|ll|ll|}
\hline
& \multicolumn{2}{c|}{Validation} & \multicolumn{2}{c|}{Testing} \\
 & Precision & Recall & Precision & Recall \\ \hline
\textbf{Edge Validity} & & & & \\
Valid     & 94.5\% & 93.8\% & 94.9\% & 91.7\%  \\
Invalid   & 79.7\%    & 81.7\% & 75.4\% & 83.7\%    \\ \hline
\textbf{Edge Category} & & & & \\
Trunk    & 62.4\%  & 65.6\% & 57.9\% & 60.9\%    \\
Support  & 89.2\%    & 62.5\% & 57.0\% & 56.4\%     \\
Leader   & 70.9\%    & 77.1\% & 91.7\% & 51.3\%    \\
Side branch & 42.1\%    & 63.6\% & 54.9\% & 63.3\%    \\
Other    & 46.7\%    & 47.3\% & 48.3\% & 66.3\%    \\ \hline
\end{tabular}
\caption{Per-classification performance results for the neural network on the edge validity and edge category tasks.}
\label{table:trainingresults}
\end{table}

The overall results of the training process are shown in Table \ref{table:trainingresultsoverall}, with accuracy defined as the number of correctly-classified edges over the total number of edges. Table \ref{table:trainingresults} shows a detailed per-category breakdown with precision and recall metrics. Letting $TP$ represent the number of true positives, $FP$ false positives, and $FN$ false negatives, precision is defined as $\frac{TP}{TP + FP}$, while recall is defined as $\frac{TP}{TP + FN}$.

Overall, the network does a good job on the edge validity task, achieving a 90\% accuracy on the testing data set. As shown through the precision/recall breakdown, the network has a more difficult time correctly identifying invalid edges as being invalid than with valid edges. This is not entirely surprising, as the human annotator's decision to label a branch edge was based off of a semi-local view of the edges using neighboring points, context which the network lacks. Furthermore, while most of the rasterized edge images were fairly clear for the human annotator to mark as being valid or invalid, there were a notable number of cases which were less clear cut, such as edges which cut closely across a corner of two branches or which were slightly off center; as such, we expect there to be some label noise which affects the performance metrics. Overall, the network performs well enough to be used for our algorithm, which is robust to small amounts of branch misclassifications.

However, for the edges which are valid, the network has a much harder time with the edge categorization task, only achieving a 59\% accuracy on the testing set. While this is better than randomly guessing, our impression was that this performance was not robust enough to be able to directly incorporate into the algorithm. Again, the underperformance at the Edge Category task is not too surprising, since the network only sees a local view of the edge and only has the position and elevation of the edge to guide it in determining what the actual edge category should be. Furthermore, an ideal classification system would be able to use the characteristics of its neighboring edges to refine its belief, e.g. an edge which is located between two edges which are confidently identified as Leaders should also probably be a Leader. Our network is unable to utilize this context due to the purely local assessment of the edge. Despite the fact that we decided not to use the predicted edge labels in our algorithm, we ultimately decided to keep them as part of the network, as it is likely that the network's performance on the edge validity task benefits from also attempting to consider how likely an edge is to belong to any given category. Future research may focus on how we can improve the ability of our network to predict the label of any given edge by considering its neighbors and global context.


\end{document}